\pdfoutput=1

\documentclass[11pt]{article}

\usepackage[preprint]{acl}

\usepackage{times}
\usepackage{latexsym}

\usepackage[T1]{fontenc}

\usepackage[utf8]{inputenc}

\usepackage{microtype}

\usepackage{inconsolata}

\usepackage{graphicx}

\title{Multimodal Needle in a Haystack: Benchmarking Long-Context Capability of Multimodal Large Language Models}

\author{
Hengyi Wang\textsuperscript{1}\thanks{Correspondence to: Hengyi Wang \textless hengyi.wang@rutgers.edu\textgreater}~,~
Haizhou Shi\textsuperscript{1},~
Shiwei Tan\textsuperscript{1},~
Weiyi Qin\textsuperscript{1},~
Wenyuan Wang\textsuperscript{1},
\\
\textbf{
Tunyu Zhang\textsuperscript{1},~
Akshay Nambi\textsuperscript{2},}~
\textbf{Tanuja Ganu\textsuperscript{2},}~
\textbf{Hao Wang\textsuperscript{1}} \\\\
\vspace{1cm}
\textsuperscript{1}{Rutgers University},
\textsuperscript{2}{Microsoft Research}
\vspace{-0.7cm}
\\
\href{https://mmneedle.github.io/}{https://mmneedle.github.io}
}

\usepackage[utf8]{inputenc} 
\usepackage[T1]{fontenc}    
\usepackage{url}            
\usepackage{booktabs}       
\usepackage{amsfonts}       
\usepackage{microtype}      

\usepackage{xcolor}         
\usepackage{tcolorbox} 
\usepackage{bm}
\tcbset{
    top=0.5mm,             
    bottom=0.5mm           
}

\usepackage{graphicx}

\def\red{\textcolor{red}}

\def\green{\textcolor{green}}

\def\0{{\bf 0}}
\def\1{{\bf 1}}

\newcommand{\tabref}[1]{Table~\ref{#1}}
\newcommand{\secref}[1]{Sec.~\ref{#1}}
\newcommand{\figref}[1]{Fig.~\ref{#1}}

\newcommand{\appref}[1]{Appendix~\ref{#1}}

\renewcommand{\hat}{\widehat}
\renewcommand{\frac}{\tfrac}

\DeclareMathAlphabet{\mymathbb}{U}{bbold}{m}{n}
\definecolor{green}{rgb}{0,0.5,0}

\def\red#1{\textcolor{red}{#1}}
\def\green#1{\textcolor{green}{#1}}

\usepackage{amsmath}
\usepackage{amssymb}
\usepackage{mathtools}
\usepackage{amsthm}
\usepackage{wrapfig}
\usepackage{booktabs}
\usepackage{algorithmicx}
\usepackage{algpseudocode}
\usepackage{enumitem}

\usepackage{multirow}
\usepackage[nice]{nicefrac}
\usepackage{enumitem}
\usepackage{tcolorbox}
\setitemize{nosep,leftmargin=18pt}

\begin{document}
\maketitle




\begin{abstract}

Multimodal Large Language Models (MLLMs) have shown significant promise in various applications, leading to broad interest from researchers and practitioners alike. However, a comprehensive evaluation of their long-context capabilities remains underexplored. To address these gaps, we introduce the MultiModal Needle-in-a-haystack (MMNeedle) benchmark, specifically designed to assess the long-context capabilities of MLLMs. Besides multi-image input, we employ image stitching to further increase the input context length, and develop a protocol to automatically generate labels for sub-image level retrieval. Essentially, MMNeedle evaluates MLLMs by stress-testing their capability to locate a target sub-image (needle) within a set of images (haystack) based on textual instructions and descriptions of image contents. This setup necessitates an advanced understanding of extensive visual contexts and effective information retrieval within long-context image inputs. With this benchmark, we evaluate state-of-the-art MLLMs, encompassing both API-based and open-source models. The findings reveal that GPT-4o consistently surpasses other models in long-context scenarios, but suffers from \mbox{hallucination} problems in negative samples, i.e., when needles are not in the haystacks. Our comprehensive long-context evaluation of MLLMs also sheds lights on the considerable performance gap between API-based and open-source models. 
{All the code, data, and instructions required to reproduce the main results are available at~\href{https://github.com/Wang-ML-Lab/multimodal-needle-in-a-haystack}{https://github.com/Wang-ML-Lab/multimodal-needle-in-a-haystack}.}

\end{abstract}

\section{Introduction}
Recent breakthroughs in multimodal large language models (MLLMs) have enabled a wide range of applications, spanning from visual question answering to cross-modal retrieval~\cite{yue2023mmmu,ying2024mmtbench}. To evaluate the capabilities and limitations of MLLMs, various benchmarks have been proposed, focusing on challenges such as reasoning~\cite{yue2023mmmu,padlewski2024vibe,lu2023mathvista}, perception~\cite{fu2024blink,yu2023mmvet}, and hallucination~\cite{guan2023hallusionbench}. 


Despite significant progress, the evaluation of MLLMs for long-context understanding has been lagging. 
Current evaluation methods and benchmarks~\cite{yue2023mmmu,ying2024mmtbench,liu2023mmbench,padlewski2024vibe,fu2024blink,yu2023mmvet,chen2024mmstar,fu2024mme,lu2023mathvista,reid2024gemini} either (1)~assume the use of single or limited images as inputs, failing to stress-test MLLMs' long-context capabilities or (2)~only contain a limited numbers of data points (referred to as ``samples'' in this paper), lacking in statistical significance and therefore often rendering the evaluation inconclusive. These gaps limit the development of MLLMs capable of effectively handling long-context hybrid-modality inputs, {which is} crucial for broader applications. 

\begin{figure*}[!t]
        \centering
        \includegraphics[width = 0.9999\textwidth]
        {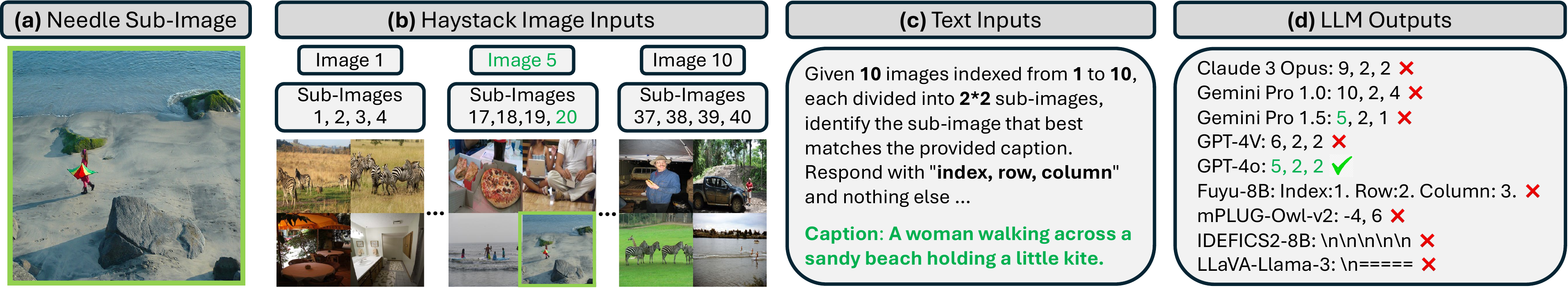}
        \vskip -0.1cm
        \caption[width = 0.8\linewidth]{MMNeedle evaluation overview. Correct answers are marked with \green{\emph{checkmark}{~($\checkmark$)}}, while the incorrect answers are marked with \red{\emph{cross}{~($\bm{\times}$})}. Our evaluation setup involves the following key components: \textbf{(a) Needle Sub-Image:} The needle sub-image to be retrieved based on the given caption. \textbf{(b) Haystack Image Inputs:} The long-context visual inputs consist of $M$ images, each stitched from $N\times N$ sub-images. \textbf{(c) Text Inputs (Instructions and Caption):} Detailed instructions to MLLMs, followed by a \green{caption} describing the needle, i.e.,  \green{sub-image $20$}. See~\secref{sec:instruction} for {MMNeedle's} complete instructions. \textbf{(d) LLM Outputs:} The answers from different MLLMs, indicating their ability to accurately locate the needle in the haystack based on the given caption. The expected output is composed of the model's identification of the index, row, and column of the matching sub-image. The results showcase the comparative performance of various models: GPT-4o correctly predicts the exact location of the needle; Gemini Pro 1.5 only correctly predicts the image index of the needle; other API models predict incorrect locations; open-source models often output with wrong formats.}
        \label{fig:overview}
        \vskip -0.2cm
\end{figure*}

To bridge this gap, we introduce the MultiModal Needle-in-a-haystack (MMNeedle) benchmark to comprehensively evaluate the long-context capabilities of MLLMs. \figref{fig:overview} shows a simple example: {The MLLMs are presented with a \emph{haystack} of images, consisting of $M=10$ images, each containing $N\times N=2\times 2=4$ sub-images (see Figure 1(b)). Additionally, a caption is provided for one of the sub-images in the \emph{haystack}, as shown in \green{green} text in Figure 1(c). The goal of the MLLMs is to identify the \emph{needle}, namely the sub-image highlighted in the \green{green} box in Figure 1(a), which corresponds to the caption.}

By using advanced techniques, such as image stitching to increase input context length, we assess MLLMs' ability to locate a target sub-image (needle) within a large set of images (haystack) based on textual instructions, i.e., instructions with the target caption in~\figref{fig:overview}(c). The highlights of our MMNeedle benchmark include:
\begin{itemize}
\item \textbf{Comprehensive Dataset.} Our {dataset} ensures sufficient samples for each setting, {with a total number of} $40{,}000$ images, $560{,}000$ captions, and $280{,}000$ needle-haystack pairs. 
\item \textbf{Diverse Settings.} {Our benchmark covers {diverse settings} with \emph{varying context lengths}, \emph{single and multiple needles}, as well as \emph{positive and negative samples}, among others} (details in~\secref{sec:methods}). 
\item \textbf{Coarse-to-Fine Evaluation Metrics.} We establish a set of evaluation metrics, including ``existence accuracy'', ``index accuracy'', and ``exact accuracy'', to holistically evaluate MLLM at the sequence-, image-, and sub-image- levels (details in~\secref{sec:metrics}). 
\item \textbf{Wide Coverage.} Our evaluation covers both state-of-the-art API-based and state-of-the-art open-source MLLMs, shedding light on their long-context capabilities.  
\end{itemize}


Our findings underscore a considerable performance gap between models and reveal the hallucination problem in state-of-the-art MLLMs through negative samples. For example, we find that 
(1) there is still a large performance gap between state-of-the-art API-based and state-of-the-art open-source models, 
(2) accuracy drops significantly with more images in the haystacks, 
even for state-of-the-art API-based MLLMs such as Claude~3~Opus and Gemini 1.0 Pro, and 
(3) all models (including Claude~3~Opus, Gemini 1.5 Pro, and GPT-4V) perform poorly in MMNeedle settings with sub-images (e.g., $N \times N = 2\times 2 = 4$ sub-images in~\figref{fig:overview}); this is true even for the best model, GPT-4o, whose accuracy drops from $97.00\%$ for $M=10$ images without sub-images (i.e., equivalent to $10$ images in the haystack) to $26.90\%$ for $M=10$ images with $N \times N = 4\times 4=16$ sub-images for each image (equivalent to $160$ images in the haystack). 
See~\figref{fig:vis} and more results in~\secref{sec:exp}.


\section{Related Work}

Existing benchmarks for MLLMs mainly focus on limited image inputs, such as reasoning~\cite{yue2023mmmu,padlewski2024vibe,lu2023mathvista,song2024milebench}, perception~\cite{fu2024blink,yu2023mmvet}, hallucination~\cite{guan2023hallusionbench}, where the answers are based on either single or only a handful of images. They are therefore not suitable for evaluating MLLMs' long-context capability for visual inputs. Recent work~\cite{fu2024data,kuratov2024search,levy2024same,zhao2024longagent} on LLMs employs the needle-in-a-haystack test~\cite{kamradt2023needle} to evaluate the long-context capability of large language models (LLMs), where the LLM is expected to answer the question by finding the corresponding information among a long irrelevant corpus as context. However, these datasets and benchmarks are not applicable for the multimodal setting. 
Google's technical report~\cite{reid2024gemini} has showcased Gemini 1.5 Pro's capability of finding the needle in an audio or video haystack. However, its evaluation (1) involves only \emph{one single sample} rather than a complete dataset, obviously lacking statistical significance and therefore rendering the evaluation inconclusive~\footnote{Our MMNeedle results show that Gemini 1.5 Pro's performance does drop a lot with long contexts, especially with multiple sub-images in the same image.}, and (2) does not involve a large set of unrelated images, which is the focus of MMNeedle. 
There is also work on the retrieval capability of small objects in a single large image~\cite{pawlowski2019needles} or retrieval from large external image datasets~\cite{brogan2019needle}, but none of them are concerned with in-context image retrieval, particularly for long-context multimodal evaluation. 

In contrast to existing benchmarks, our MMNeedle benchmark includes a dataset of $40{,}000$ images, $560{,}000$ captions, and $280{,}000$ needle-haystack pairs (more details in~\secref{sec:methods}), 
rather than only one (or a handful of) needle-haystack pair(s)~\cite{kamradt2023needle,reid2024gemini}. MMNeedle also includes a diverse set of metrics and evaluation protocols, covering different numbers of needle sub-images and needle sub-images. These differences set MMNeedle apart from existing benchmarks and are essential to evaluate MLLMs' long-context capability comprehensively. 


\section{MultiModal Needle in a Haystack (MMNeedle)}\label{sec:methods}

In this section, we introduce our MultiModal Needle-in-a-haystack (MMNeedle) benchmark. 

\subsection{Overview} \label{sec:overview}



\textbf{Problem Setting.} 
\figref{fig:overview} provides an overview of our evaluation setup with a randomly selected example from our MMNeedle dataset (details in~\secref{sec:construct_long_context}). 
The MLLM is given (1) an image \emph{haystack}, i.e., a sequence of $M$ images, ($M=10$ in~\figref{fig:overview}), with each image containing $N\times N$ sub-images ($N=2$ in~\figref{fig:overview}), and (2) a caption for one of the sub-images, shown as \green{green} text in \figref{fig:overview}(c). The MLLM's goal is then prompted to find the \emph{needle}, i.e., the sub-image which the caption describes. 
Note that our evaluation setup can be naturally applied for video-based inputs {by extracting images from individual frames}, which would be interesting future work. 



\textbf{Evaluation Goals.} 
As illustrated in~\figref{fig:overview}, our MMNeedle aims to evaluate the MLLMs' \emph{three key capabilities} within one forward pass: (1) understanding the semantics of both visual and textual inputs, 
(2) retrieving the sub-image (needle) from long-context images (haystack), and (3) understanding and following the instructions~\cite{xia2024fofo} to output the location of the sub-image (needle) in the correct format.

\subsection{MMNeedle Dataset}\label{sec:construct_long_context}

\textbf{Constructing Long Context.}
To evaluate the long-context capability of MLLMs, we extend the context length of visual inputs in the following two aspects:
\begin{itemize}
    \item \textbf{More Images:} We increase the number of images in the inputs for MLLMs to extend the visual context length. Specifically, we use two different numbers of images $M$ in the prompt, i.e., $M=1$ or $M=10$. Note that {\textbf{we choose $M=10$ because it is the largest number of input images that GPT-4V/GPT-4o can support (see~\tabref{tab:maximum_images_brief} and~\appref{app:dataset}).}} 
    \item \textbf{Image Stitching:} We stitch small images into a single large image {as the input}. Specifically, we use $N\times N$ sub-images ($N \in \{1, 2, 4, 8\}$) to compose a stitched image with $N$ rows and $N$ columns, each combination of row and column indices {(r,c)} corresponding to a sub-image. {\figref{fig:overview}(b) shows an example of $2\times 2$ stitching{, with $4$ sub-images in $1$ stitched image.}} 
    
\end{itemize}



\begin{table}[!t]
    \centering
    \vskip 0.1cm
    \caption{Maximum numbers of images per request for \underline{Az}ure \underline{GPT-4}V/o , \underline{Op}enAI \underline{GPT-4}V/o, \underline{Claude}, and \underline{Gemini}. "*" indicates that the OpenAI GPT-4V/o API supports at most $10$ images with high quality. Other numbers are \emph{hard} limits. See~\appref{app:dataset} for details.}
    \label{tab:maximum_images_brief}
    \vskip -0.2 cm
    \resizebox{0.48\textwidth}{!}{
    \begin{tabular}{lcccc}
        \toprule
        \textbf{Model} & GPT-4 (Az.) & GPT-4 (Op.) & Claude &Gemini\\
        \midrule
        \textbf{Limit}& 10 
         & 10$^*$ 
         & 20 
        & 16 \\
        \bottomrule
    \end{tabular}}
    \vskip -0.6 cm
\end{table}
\textbf{Purpose of Image Stitching.} 
{The purpose of \emph{image stitching} is to: (1) Extend the effective context length. For example, stitching $M=10$ images, each with $N\times N=8\times 8$ sub-images, results in a long context of $640$ sub-images. This setup tests MLLMs' long-context capabilities. (2) Test MLLMs' localization capability by requiring them to pinpoint sub-images within a large image based on specific captions.} {For details, see~\appref{app:dataset}.}

Combining both dimensions provides comprehensive settings for our evaluation: $(M, N) = (1, 2), (1, 4), (1, 8), (10, 1), (10, 2), (10, 4), (10, 8)$. Note that $(M, N) = (1, 1)$ is excluded, as finding an image within a single image is trivial. {Note that MMNeedle covers typical, real-world MLLM use-cases. Specifically, single, complete images correspond to our setting with the number of images $M=10$ and the stitch size $N\times N = 1\times 1$.}


\textbf{Single-Needle Setting, Multi-Needle Setting, and the Number of Needles $K$.} 
We also extend the single-needle setting above, i.e., the number of needles (and associated captions) per query $K=1$, to a multi-needle setting, where there are $K > 1$ needles.

\textbf{Image Data.} 
In this paper, we use the MS COCO 2014 validation set~\cite{lin2014microsoft} as our source dataset for constructing our MMNeedle dataset. Note that our data construction approach is agnostic to the dataset and can be applied to any dataset containing images with paired captions that describe the content of the images. 
We resize each original image from the MS COCO 2014 validation set to $256 \times 256$ pixels before stitching them into a larger image. 
The image resolution of $256$ pixels is chosen to ensure sufficient image quality; our preliminary studies show that humans (and MLLMs) cannot effectively recognize MS COCO images with resolution lower than $256$ ({see examples in~\figref{fig:overview} and more in~\appref{app:dataset}}). 
We then stitch these sub-images using stitching sizes of $1 \times 1$, $2 \times 2$, $4 \times 4$, and $8 \times 8$, leading to larger images with resolutions of $256 \times 256$, $512 \times 512$, $1024 \times 1024$, and $2048 \times 2048$, respectively. 
{Given that Claude 3 supports a maximum resolution of $1092 \times 1092$ pixels and GPT-4 (including GPT-4V and GPT-4o) supports a maximum resolution of 2000 pixels for the long side of an image, we have chosen 2048 pixels as the maximum resolution for our stitched images. Note that these models will resize images that exceed their respective size limits.} 

\subsection{Dataset Construction: Automated Sampling} \label{sec:dataset_sample}

\textbf{Positive and Negative Samples.} 
Our dataset is divided into (1) positive samples, where a sub-image (needle) exists in the context (haystack) to match the given caption, and (2) negative samples, where no sub-image (needle) exists in the context that can match the given caption. To construct the dataset with balanced data distribution, we generate 5000 samples each for positive and negative samples for each $(M,N,K)$ combination, leading to $280{,}000$ needle-haystack pairs in total. 

\textbf{Sampling Process.}
Specifically, we construct our dataset with the following sampling process:
\begin{itemize}[leftmargin=18pt]
    \item\textbf{Step 1: Sampling Single-Image Haystacks.} For each stitch size $N\in \{1, 2, 4, 8\}$, we first construct $10{,}000$ stitched images, with each sub-image randomly sampled from the MS COCO validation dataset (ensuring each stitched image has no repetitive sub-images). 
    These $10{,}000$ stitched images 
    directly constitute the haystacks for stitching size $N$ in the $M=1$ setting. 
    \item\textbf{Step 2: Sampling Multi-Image Haystacks.} For each stitch size $N\in \{1, 2, 4, 8\}$ in the $M=10$ setting, we sample 10 different images as a haystack from the $10{,}000$ stitched images constructed in Step 1. We sample $10{,}000$ such haystacks for stitching size $N$ (ensuring each haystack has no repetitive stitched images). 
    \item\textbf{Step 3: Generating Positive Samples.} 
    We sample a sub-image as a needle from a unique haystack (i.e., $M \times N \times N$ sub-images) in Step 1 or Step 2, obtain its associated caption MS COCO annotations, and use this caption as the query in our MMNeedle evaluation {(see~\figref{fig:overview})}. 
    We repeat this process for $K$ times in multi-needle settings, where $K=2$ or $K=5$ (ensuring each needle is a unique sub-image). 
    This process ensures that the needles are \emph{inside} the haystack. 
    \item\textbf{Step 4: Generating Negative Samples.} 
    From the MS COCO 2014 validation set, we sample an image outside the haystack in Step 1 or Step 2 and use the image as the needle for a negative sample. We also obtain the needle's associated caption from MS COCO annotations and use it as the query in our MMNeedle evaluation. We repeat this process for $K$ times in multi-needle settings, where $K=2$ or $K=5$ (ensuring each needle refers to a unique sub-image). 
    This ensures that the needles are \emph{outside} the haystack. 
\end{itemize}

With the process above, we construct $5{,}000$ \emph{positive} and $5{,}000$ \emph{negative} samples for each setting $(M, N, K)$, where $M\in \{1,10\}$, $N\in \{1, 2, 4, 8\}$, and $K\in \{1, 2, 5\}$. 

\subsection{Evaluation Metrics}\label{sec:metrics}

As mentioned in the previous sections, there are two ``axes'' for different settings in our MMNeedle evaluation: (1) the number of input images $M$, which indicates how many images are passed as inputs to an MLLM, and (2) the stitching size $N$, where $N$ is the number of total columns/rows of sub-images (where $N=1$ means that each input image is the original image from the MS COCO 2014 validation set, otherwise, it is $N \times N$ images stitched as one). Increasing each of these axes adds difficulty to MLLMs due to the increased context length, i.e., the {haystack size}. 
We propose and use the following evaluation metrics:

\textbf{Single Needle.} For the single-needle setting, we define three different metrics to evaluate as follows:
\begin{itemize}
\item \textbf{Existence Accuracy} is the proportion of samples in which the model correctly predicts \emph{whether the needle exists} {in the input image sequence}. 
\item \textbf{Index Accuracy} is the proportion of samples where the model correctly predicts the index $m\in\{1,\dots,M\}$ of the stitched image containing the needle (e.g., $m=5$ in~\figref{fig:overview}). 
\item \textbf{Exact Accuracy} (\emph{success rate} of the needle retrieval~\cite{reid2024gemini}) is the proportion of samples where the model correctly predicts the needle sub-image's location, i.e., index $m$, row $r$ and column $c$. 
\end{itemize}

\textbf{Multiple Needles.} We use similar metrics for the multi-needle setting (details in~\appref{app:protocol}). 


\textbf{Coarse-to-Fine Evaluation.} 
From the definitions, we can see that these accuracies satisfy the relation $\text{``Existence Accuracy''} \ge \text{``Index Accuracy''} \ge \text{``Exact Accuracy''}$ for a given model and evaluation setting $(M, N, K)$. This indicates a coarse-to-fine evaluation using our devised metrics. 

\textbf{Automated Evaluation Protocol.} 
We design an automated evaluation protocol for the defined three metrics as follows:
\begin{itemize}
    \item \textbf{Ground Truth Format.} (1) For each positive sample, i.e., the needle sub-image is in the context, the ground-truth output is {``$m, r, c$''} that describes the location of the needle, where $m$ is the image index ($m \in {1,...,M}$), and $r, c$ are the row and column of the sub-image (needle) in image $m$, respectively ($r, c \in {1,...,N}$). (2) For each negative sample, i.e., no needle sub-image is in the context, the ground-truth output is \textbf{``-1''}, indicating the needle does not exist. 
    The multi-needle setting uses a similar format (details in~\appref{app:protocol}). 
    \item \textbf{Existence Accuracy} is measured by whether the MLLM outputs ``-1'' 
    (in multi-needle settings, we match ``-1'' for all the needles, separated by ``;'', or alternatively just one ``-1''). Specifically, for positive samples (targets exist), the existence accuracy is the proportion of samples where the MLLM does not predict ``-1'', and for negative samples (targets do not exist), the existence accuracy is the proportion of of samples where the MLLM predicts ``-1'' (see~\secref{sec:acc_results} for details).
    \item \textbf{Index Accuracy} is measured by whether the image index $\hat{m}$ predicted by MLLM matches the ground truth $m$. For multi-needle settings, predictions are considered correct only if the MLLM predicts the correct $m$ for \emph{all needles}. Note that even for the $M=1$ settings, the index accuracy may not be perfect (100\%), because the model can fail to output the only image index ``1''. Therefore, we also evaluate the index accuracy of different models in the $M=1$ settings (see~\secref{sec:acc_results} for details).
    \item \textbf{Exact Accuracy} is measured by whether the tuple $(\hat{m}, \hat{r}, \hat{c})$ predicted by MLLM matches the ground truth $(m, r, c)$. For multi-needle test, predictions are considered correct only if the MLLM predicts the correct $(m, r, c)$ for all needles. 
\end{itemize}

 

\begin{figure*}[!t]
        \centering
        \includegraphics[width = 0.99\textwidth]
        {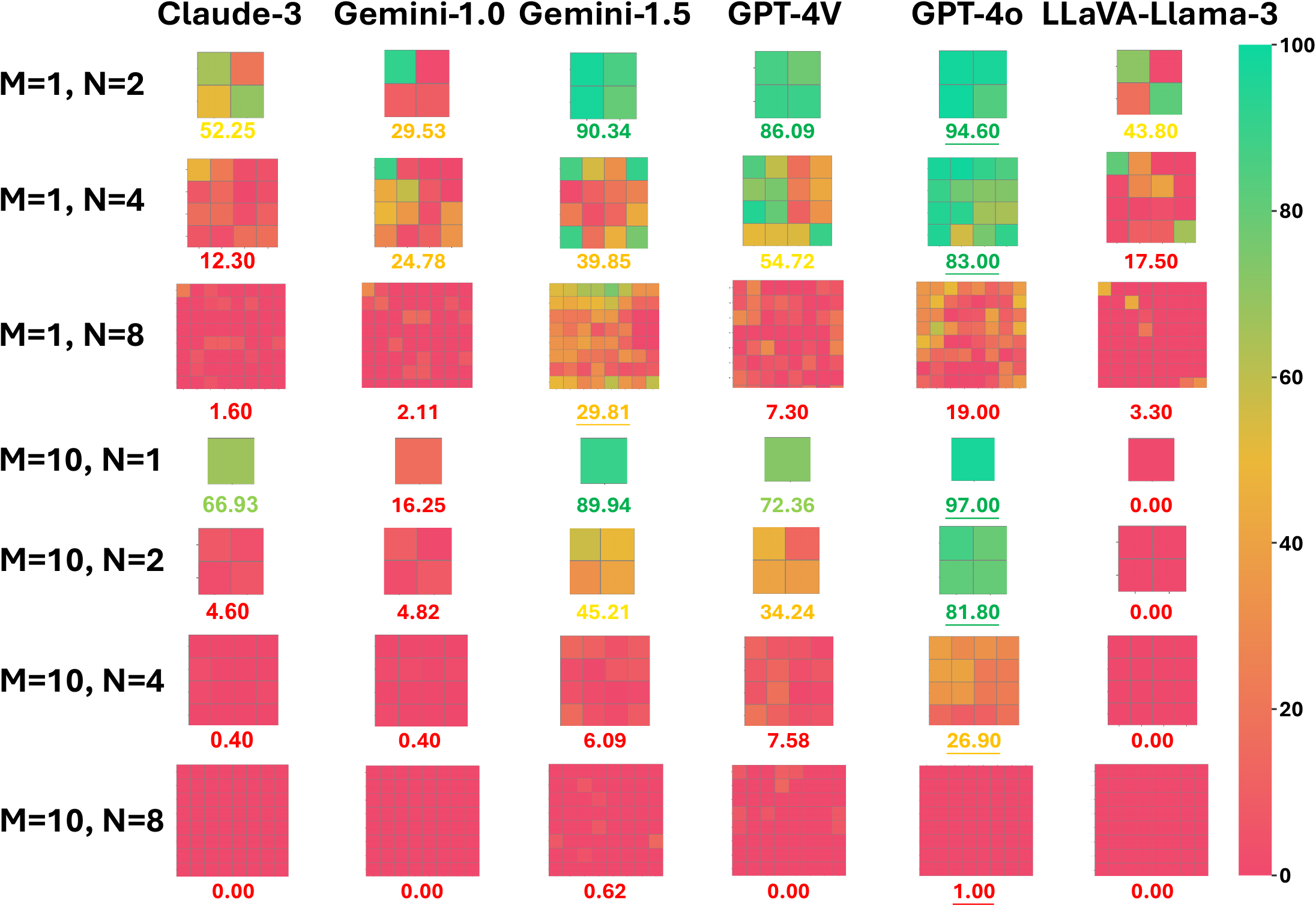}
        \vskip -0.1cm
        \caption[width = 0.8\linewidth]{MMNeedle evaluation performance comparison (Claude-3 refers to Claude~3~Opus, and Gemini-1.0/1.5 refers to Gemini Pro 1.0/1.5). The x-axis shows the results of different models, and the y-axis shows the results on various input image number $M$ and stitching size $N$. For each row, i.e., setting $(M,N)$, we show the average accuracy ($\%$) of each model. For each stitched image, the color of row $r$, the column $c$ indicates the accuracy of predicting the exact position for samples with the ``needle'' sub-image in position $(r,c)$ of the stitched image. For the $M=10$ setting, we show the average accuracy of each location $(r,c)$ over 10 images. A \red{\emph{redder}} cell indicates lower accuracy, while a \green{\emph{greener}} cell indicates higher accuracy. 
        The best result for each row is marked with \underline{underlining}.   
        }
        \label{fig:vis}
        \vskip -0.3cm
\end{figure*}

\section{Experiments}\label{sec:exp}
In this section, we describe the evaluation results of various MLLMs on our MMNeedle dataset.


\subsection{Evaluated MLLMs}\label{sec:exp_model}
We conduct MMNeedle evaluation for both API-based models and open-source models:
\begin{itemize}
    \item \textbf{API-Based Models.} We evaluate state-of-the-art API-based MLLMs, including \textbf{Claude 3 Opus} (Feb 2024)~\cite{anthropic2023claude3}, \textbf{Gemini Pro 1.0} (Feb 2024)~\cite{team2023gemini}, \textbf{Gemini Pro 1.5} (May 2024)~\cite{reid2024gemini}, \textbf{GPT-4V} (March 2024)~\cite{achiam2023gpt4}, and \textbf{GPT-4o} (May 2024)~\cite{openai2024gpt4o}. 
    \item \textbf{Open-Source Models.} We evaluate top open-source multimodal LLMs, including \textbf{CogVLM} (CogVLM-17B/CogVLM2-Llama-3)~\cite{wang2023cogvlm}, \textbf{Fuyu-8B}~\cite{fuyu-8b}, \textbf{mPLUG-Owl-v2}~\cite{ye2023mplug}, \textbf{{InstructBLIP}} (InstructBLIP-Vicuna-13B/InstructBLIP-Flan-T5-XXL)~\cite{dai2024instructblip}, \textbf{IDEFICS2}~\cite{laurenccon2024idefics2}, and \textbf{{LLaVA-Llama-3}}~\cite{li2024llavanext-strong}. Note that CogVLM and InstructBLIP do \emph{not} support multi-image inputs; therefore, we do not test them for our multi-image ($M=10$) settings.
\end{itemize}

See~\appref{app:implement} for more details on evaluated MLLMs.

\subsection{Overview of MMNeedle Evaluation Results}
\figref{fig:vis} shows an intuitive comparison of the \emph{exact accuracy} (defined in~\secref{sec:metrics}) across advanced MLLMs in various single-needle ($K=1$) settings, including Claude~3~Opus, Gemini Pro 1.0, Gemini Pro 1.5, GPT-4V, GPT-4o, and LLaVA-Llama-3. 
Each heatmap is divided into $N\times N$ cells, where the cell at row $r$, column $c$ is marked in a color that indicates the average accuracy of the model predicting the exact location for needle sub-images at $(m, r, c)$ ($m$ is the image index of the needle). 
We highlight the following observations: 
\begin{itemize}
\item \textbf{Impact of Stitching Size $N$ and Input Image Number $M$:} 
For an MLLM (one column in~\figref{fig:vis}), if we fix the number of input images $M$, the accuracy drops quickly when increasing the stitching size $N$. This drop is more significant for $M=10$ than for $M=1$, where the accuracy drops to near zero for all models on samples with $M=10, N=8$.
\item \textbf{Capability of the API-Based Models:} For a fixed $(M, N)$ pair (one row in~\figref{fig:vis}), the performance varies significantly for different MLLMs, particularly for samples with low stitching size $N$. GPT-4o achieves the highest accuracy except for $M=1, N=8$ samples, where Gemini Pro 1.5 reaches the best performance and GPT-4o is the second-best.
\item \textbf{Capability of the Open-Source Models:} LLaVA-Llama-3, as a top open-source model, enjoys comparable performance with frontier API-based models such as Claude~3~Opus and Gemini Pro 1.0 for $M=1$ samples, while lagging behind in $M=10$ samples.
\end{itemize}
{We also analyze the error patterns. As illustrated in~\figref{fig:vis}, the models demonstrate higher accuracy when the needles are positioned in the corners of the image compared to when they are located in the center. This trend is particularly pronounced in Gemini-1.5 and LLaVA-Llama-3, in contrast to GPT-4o.} 
See~\secref{sec:acc_results} below for details and more evaluation results.

\begin{table*}[!t]
\caption{Accuracy (\%) for the $M=1$ setting. We mark the best results with \textbf{bold face}. Note that the existence accuracy is measured by whether the model outputs ``-1''. The index accuracy is not always $100\%$ because the model can fail to output the only image index ``1''.}\label{tab:img1_needle1}
\vskip -0.1cm
\centering
\resizebox{0.99\textwidth}{!}{%
\begin{tabular}{llccccccccccccccc}
\toprule
 & Stitching& \multicolumn{3}{c}{$2\times2$} & \multicolumn{3}{c}{$4\times4$} & \multicolumn{3}{c}{$8\times8$}  \\
\cmidrule(r){2-2} \cmidrule(r){3-5} \cmidrule(r){6-8} \cmidrule(r){9-11}
& Metrics & \emph{Existence} & \emph{Index} & \emph{Exact} & \emph{Existence} & \emph{Index} & \emph{Exact} & \emph{Existence} & \emph{Index} & \emph{Exact}\\
\midrule
\multirow{5}{*}{API-Based Models}&
Claude~3~Opus& 75.38&74.77&52.25&58.70&58.00&12.30&56.36&54.85&1.60\\
&Gemini Pro 1.0 &97.10&85.09&29.53	&88.42&82.88&24.78&	55.62 & 45.18 & 2.11 \\
&Gemini Pro 1.5& {99.59} & {99.38} & {90.34}&	{98.85} & {98.44} & 39.85&	96.65 & 96.65 & \bf{29.81}\\
&GPT-4V & 92.64 &92.64&86.09	&97.29&97.19&{54.72}&	{98.20}&98.20&7.30\\
&GPT-4o & 99.00&99.00&\bf{94.60}&{99.50}&{99.50}&\bf{83.00}&{99.60}&\bf{99.60}&19.00 \\
\midrule
\multirow{8}{*}{Open-Source Models}& 
CogVLM-17B& {99.90}&0.80&0.00&97.50&3.30&0.10&96.90&22.90&0.30 \\
&CogVLM2-Llama-3 & 69.10&24.60&7.30&69.90&16.40&0.90&55.90&5.30&0.10\\
&Fuyu-8B & \bf{100.00}&0.50&0.00&\bf{100.00}&0.00&0.00&\bf{100.00}&0.00&0.00\\
&mPLUG-Owl-v2 & 96.60&48.60&1.90 & 90.70&34.30&0.30&86.30&36.90&0.70\\
&InstructBLIP-Vicuna-13B &\bf{100.00}&6.90&0.00&\bf{100.00}&11.70&0.00&\bf{100.00}&32.00&0.00\\ 
&InstructBLIP-Flan-T5-XXL&\bf{100.00} & \bf{100.00} & 3.80&\bf{100.00} & \bf{100.00} &6.20&\bf{100.00} &93.00&2.20\\
&IDEFICS2-8B & 75.80&69.30&18.90& 95.80&86.00&7.80&39.60&24.50&0.90\\
&LLaVA-Llama-3 &\bf{100.00} & 93.70& 43.80 & 97.20&93.00&17.50&95.40&95.30&3.30\\
\bottomrule
\end{tabular}
}
\end{table*}
\begin{table*}[t]
\caption{Accuracy (\%) for the $M=10$ setting.  We mark the best results with \textbf{bold face}. Note that the existence accuracy is measured by whether the model outputs ``-1''.}\label{tab:img10_needle1}
\vskip  -0.1 cm
\centering
\resizebox{0.99\textwidth}{!}{%
\begin{tabular}{llcccccccccccccccccc}
\toprule
&Stitching & \multicolumn{3}{c}{$1\times1$} & \multicolumn{3}{c}{$2\times2$} & \multicolumn{3}{c}{$4\times4$} & \multicolumn{3}{c}{$8\times8$}  \\
\cmidrule(r){2-2} \cmidrule(r){3-5} \cmidrule(r){6-8} \cmidrule(r){9-11} \cmidrule(r){12-14} 
&Metrics & \emph{Existence} & \emph{Index} & \emph{Exact} & \emph{Existence} & \emph{Index} & \emph{Exact} & \emph{Existence} & \emph{Index} & \emph{Exact}& \emph{Existence} & \emph{Index} & \emph{Exact}\\ 
\midrule
\multirow{5}{*}{API-Based Models}&
Claude~3~Opus& 83.77&67.23&66.93& 66.60&9.90&4.60&64.78&6.46&0.40& {54.13}&{5.93}&{0.00}\\
&Gemini Pro 1.0 & 83.66 & 33.90 & 16.25&	81.63 & 10.74 &4.82	&58.92 & 4.81 & 0.40&	18.11 & 1.61 & 0.00\\
&Gemini Pro 1.5& {97.08} & {90.04} & {89.94}&	{98.84}&  {53.42}& {45.21}&	96.17& 17.26& 6.09&	89.02& 9.86& {0.62}\\
&GPT-4V & 95.11&75.59&72.36&98.32&52.10&34.24&	{99.80}&{24.87}&{7.58}&	{99.50}&{10.57}&0.00\\
&GPT-4o& {99.00}&\bf{97.00}&\bf{97.00} & {99.60}&\bf{87.20}&\bf{81.80}&\bf{100.00}&\bf{45.00}&\bf{26.90}&{99.80}&\bf{17.80}&\bf{1.00}\\
\midrule
\multirow{4}{*}{Open-Source Models}&
Fuyu-8B & \bf{100.00}&0.00&0.00&\bf{100.00}&0.00&0.00&\bf{100.00}&0.00&0.00&\bf{100.00}&0.00&0.00\\
&mPLUG-Owl-v2 & 15.90& 5.60&0.40 & 70.10&5.20&0.10 &88.50&8.10&0.00&86.10&6.30&0.00\\
&IDEFICS2-8B &71.10 &0.30&0.00&93.80&0.70&0.00&99.60&6.40&0.00&96.60&2.40&0.00\\
&LLaVA-Llama-3 &\bf{100.00}&0.20&0.00&\bf{100.00}&0.10&0.00&\bf{100.00}&0.00&0.00&\bf{100.00}&0.00&0.00 \\
\bottomrule
\end{tabular}%
}
\vskip -0.2cm
\end{table*}
\subsection{Detailed Results of the Three Defined Metrics}\label{sec:acc_results}
In this section, we discuss the results of the MMNeedle evaluation in various settings of $(M,N,K)$ across three metrics: \textbf{Existence}, \textbf{Index}, and \textbf{Exact Accuracy}, as defined in~\secref{sec:metrics}. More results are available in~\appref{app:results}. 

\textbf{Results on Single-Image Samples ($M=1$).}
\tabref{tab:img1_needle1} shows the accuracy on samples in the $M = 1$ setting, with three different stitching scenarios (i.e., $N\times N$ as $2\times 2$, $4\times 4$, and $8\times 8$). GPT-4o achieves the highest exact accuracy $94.60\%$ and $83.00\%$ for the $2\times 2$ and $4\times 4$ stitching, respectively, while Gemini Pro 1.5 achieves the highest exact accuracy, $29.81\%$, for the $8\times 8$ stitching. 
Among open-source models, LLaVA-Llama-3 performs well in simpler stitching settings, outperforming Gemini Pro 1.0 by $14.27\%$ on $2\times 2$ stitching, and Claude~3~Opus by $5.20\%$ on $4\times 4$ stitching. The results highlight that while open-source models can match or exceed API-based models in simpler contexts or metrics, they generally lag behind in more complex stitching scenarios.

\textbf{Results on Multi-Image Samples ($M>1$).}
\tabref{tab:img10_needle1} extends our evaluation to multi-image samples, i.e., $M = 10$. It shows that GPT-4o consistently performs best in terms of index/exact accuracy for all stitching sizes, outperforming other models' exact accuracy by at least $7.06\%$, $36.59\%$, $19.32\%$, and $0.38\%$ on $1\times 1$, $2\times 2$, $4\times 4$, and $8\times 8$ stitching, respectively. These results indicate stronger long-context capability of GPT-4o for multi-image samples compared to other state-of-the-art models, such as GPT-4V and Claude~3~Opus. In contrast, open-source models only achieve near-zero exact accuracy in all stitching sizes. {Note that from $1\times 1$ to $4\times 4$ stitching, GPT-4o's exact accuracy drops rapidly from $97.00\%$ to $26.90\%$, while its index accuracy drops from $97.00\%$ to $45.00\%$}; this shows that even the best performing MLLM struggles in long-context needle test, verifying the effectiveness of both our coarse-to-fine metrics and MMNeedle's dataset in stress-testing MLLMs. 

\begin{table*}[!t]
\caption{Accuracy (\%) for samples with $M=1$ in the 2-needle setting. We mark the best results with \textbf{bold face}. Existence accuracy is measured by whether the model outputs ``-1'' for \emph{all} the needles. Index accuracy is not always $100\%$ because models can fail to output the only image index ``1''.}\label{tab:img1_needle2}
\vskip -0.1cm
\centering
\resizebox{0.99\textwidth}{!}{%
\begin{tabular}{llccccccccccccccc}
\toprule
 & Stitching& \multicolumn{3}{c}{$2\times2$} & \multicolumn{3}{c}{$4\times4$} & \multicolumn{3}{c}{$8\times8$}  \\
 \cmidrule(r){2-2} \cmidrule(r){3-5} \cmidrule(r){6-8} \cmidrule(r){9-11}
& Metrics & \emph{Existence} & \emph{Index} & \emph{Exact} & \emph{Existence} & \emph{Index} & \emph{Exact} & \emph{Existence} & \emph{Index} & \emph{Exact}\\
\midrule
\multirow{5}{*}{API-Based Models}&
Claude~3~Opus & \bf{100.00}&66.00&32.00&97.00&31.00&1.00&98.00&25.00&0.00\\
&Gemini Pro 1.0 &\bf{100.00}&79.80&9.09&95.00&50.00&2.00&68.00&11.00&0.00\\
&Gemini Pro 1.5& \bf{100.00}&\bf{94.95}&\bf{87.88}&\bf{100.00}&84.00&22.00&98.00&80.00&\bf{6.00}\\
&GPT-4V &\bf{100.00}&90.72&71.13&\bf{100.00}&\bf{95.00}&{34.00}&\bf{100.00}&\bf{93.41}&1.10\\
&GPT-4o& \bf{100.00}&84.00&76.00&\bf{100.00}&84.00&\bf{57.00}&\bf{100.00}&78.00&2.00\\
\midrule
\multirow{8}{*}{Open-Source Models}&
CogVLM-17B& \bf{100.00}& 0.00 &0.00 & \bf{100.00}& 0.00 &0.00& \bf{100.00}& 0.00 &0.00\\
&CogVLM2-Llama-3& \bf{100.00}& 0.00 &0.00 & \bf{100.00}& 0.00 &0.00& \bf{100.00}& 0.00 &0.00\\
&Fuyu-8B& \bf{100.00}& 0.00 &0.00 & \bf{100.00}& 0.00 &0.00 & \bf{100.00}& 0.00 &0.00\\
&mPLUG-Owl-v2 &98.00&0.00&0.00 &94.00&2.00&0.00&96.00&3.00&0.00\\
&InstructBLIP-Vicuna-13B & \bf{100.00}& 0.00 &0.00 & \bf{100.00}& 0.00 &0.00& \bf{100.00}& 0.00 &0.00\\
&InstructBLIP-Flan-T5-XXL & \bf{100.00}& 17.00 &1.00 & \bf{100.00}& 0.00 &0.00& \bf{100.00}& 0.00 &0.00\\
&IDEFICS2-8B & \bf{100.00}& 0.00 &0.00 & \bf{100.00}& 0.00 &0.00& \bf{100.00}& 0.00 &0.00\\
&LLaVA-Llama-3  & \bf{100.00}& 0.00 &0.00 & \bf{100.00}& 2.00 &0.00& \bf{100.00}& 12.00 &0.00\\
\bottomrule
\end{tabular}%
}
\vskip -0.2cm
\end{table*}
\begin{table}[!t]
\caption{Existence Accuracy (\%) for the negative samples (the ground truth is ``-1''). We mark the best results with \textbf{bold face}. Note that the existence accuracy is measured by whether the model outputs ``-1''. 
{``-''} means that the models do not support multi-image inputs. 
}\label{tab:neg_needle1}
 \vskip  -0.1 cm
\centering
\resizebox{0.48\textwidth}{!}{%
\begin{tabular}{lccccccc}
\toprule
Stitching & \multicolumn{1}{c}{$1\times1$} & \multicolumn{2}{c}{$2\times2$} & \multicolumn{2}{c}{$4\times4$} & \multicolumn{2}{c}{$8\times8$}  \\
\cmidrule(r){1-1} \cmidrule(r){2-2} \cmidrule(r){3-4} \cmidrule(r){5-6} \cmidrule(r){7-8} 
Context & 10 imgs & 1 img& 10 imgs& 1 img& 10 imgs &1 img & 10 imgs \\ 
\midrule
{API-Based Models}\\
\midrule
Claude~3~Opus& 81.78& 77.88&54.10 & \bf{67.03}&38.38 & 51.10&53.38\\ 
Gemini Pro 1.0 & 90.60 &89.67 &	\bf{67.14}& 64.73 &	\bf{56.00} &	{57.27}&\bf{87.13}\\
Gemini Pro 1.5& \bf{92.23}&87.70&	54.56 & {65.88}&	18.77&33.75 &	17.50\\
GPT-4V & 90.57	& \bf{92.98}& 36.01&52.70 &	0.71 & 3.40&	0.10 \\
GPT-4o& 89.40&91.90 &34.80&61.60&1.30&3.10&0.20\\
\midrule 
Open-Source Models\\
\midrule
CogVLM-17B& -&3.80&-&3.50&-&2.50&-  \\ 
CogVLM2-Llama-3& -&90.30&-&65.50&-&52.70&-  \\ 
Fuyu-8B & 0.00& 0.00& 0.00& 0.00& 0.00& 0.00& 0.00 \\
mPLUG-Owl-v2 & 91.70& 36.00&35.60&16.20&12.70&12.70&13.40\\
InstructBLIP-Vicuna-13B & -& 0.00& -& 0.00& -& 0.00& -\\
InstructBLIP-Flan-T5-XXL & -& 0.00& -& 0.00& -& 0.00& -\\
IDEFICS2-8B & 30.80& 89.40& 6.90& 55.70& 0.60& \bf{62.00}& 3.10 \\
LLaVA-Llama-3 & 0.00& 11.10& 0.00& 7.40& 0.00& 5.90& 0.00 \\
\bottomrule
\end{tabular}%
}
\vskip -0.7cm
\end{table}

\textbf{Results on Multi-Needle Samples ($K>1$).}
\tabref{tab:img1_needle2} shows the results of different models on multi-needle samples, i.e., the number of needles $K=2$. Gemini Pro 1.5 achieves the highest exact accuracy $87.88\%$ on $2\times 2$ samples, and GPT-4o achieves the highest exact accuracy $57.00\%$ on $4\times 4$ samples. In contrast, the exact accuracy of open-source models is close to zero for all stitching sizes. These results indicate a large gap between the API-based and the open-source models. 
See~\appref{app:results} for more results and analysis on multi-needle samples ($K=2$ or $K=5$).

\textbf{Results on Negative Samples.}
\tabref{tab:neg_needle1} shows the existence accuracy (defined in~\secref{sec:metrics}) for negative samples (defined in~\secref{sec:dataset_sample}). For API-based models, Claude~3~Opus and Gemini Pro 1.0 
perform well 
across different configurations, suggesting robustness in handling varied context-length for the negative samples. On the other hand, GPT-4V and GPT-4o achieve inferior accuracy on more complex settings, including multi-image inputs ($M=10$) and/or large stitching size ($N=4$ or $N=8$). 
These results reveal that: (1) Even top API-based models severely suffer from hallucination; they incorrectly believe the needle exists in the haystack when it does not. (2) API-based models with \emph{stronger} needle-retrieval performance, e.g., GPT-4o, tend to \emph{suffer more from hallucination}. 

The performance of open-source models varies significantly, with some generally underperforming compared to API-based models (e.g., CogVLM-17B, Fuyu-8B, InstructBLIP and LLaVA-Llama-3), while others demonstrate high existence accuracy (e.g., CogVLM2-Llama-3, mPLUG-Owl-v2, IDEFICS2-8B). Notably, IDEFICS2-8B achieves the highest accuracy of $62.00\%$ on $M=1, N=8$ samples, indicating a low level of hallucination in this setting.

\textbf{{Summary.}}
These results show that our existence, index, and exact accuracy are designed to differentiate the model capabilities across various settings while also facilitating a transition from easier to more challenging tasks.

For example, we demonstrate that various metrics highlight the long-context capabilities of models under different settings:
\begin{itemize}
    \item \emph{Exact Accuracy:} In~\tabref{tab:img1_needle1}, where the number of input images $M=1$, we focus on evaluating exact accuracy, which measures whether the model correctly predicts both the row and column of the needle.
    \item \emph{Index Accuracy:} In~\tabref{tab:img10_needle1}, where the number of input images $M=10$, we emphasize index accuracy, assessing whether the model correctly identifies the image index within the image haystack. Together with {Exact Accuracy}, it is crucial for evaluating whether an MLLM can understand images and sub-images in the long-context scenario. 
    \item \emph{Existence Accuracy:} In~\tabref{tab:img1_needle2}, where negative samples are introduced, we evaluate existence accuracy, which reflects whether the model correctly determines that the needle is not present in the haystack. This is particularly relevant for benchmarking \textbf{hallucination} in MLLMs. 

\end{itemize}
These analyses underscore the different use cases and the necessity of our coarse-to-fine metrics. 

\begin{figure}[h]
        \centering
        \includegraphics[width = 0.48\textwidth]
        {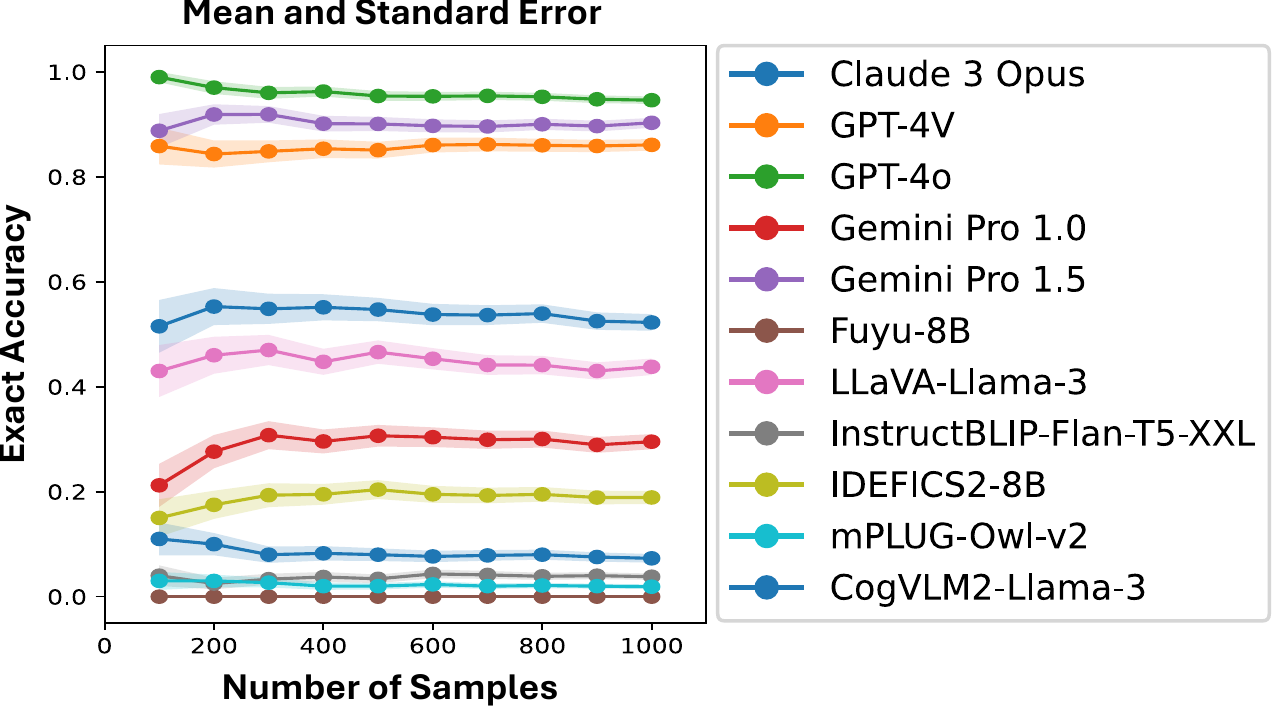}
        \vskip -0.2cm
        \caption[width = 0.8\linewidth]{Exact Accuracy of Models on Varying Sample Sizes in the $M=1, N=2$ Setting.}
        \label{fig:1img_stat}
        \vskip -0.5cm
\end{figure}

\subsection{Statistical Significance}\label{sec:stat}


\figref{fig:1img_stat} shows the results of our hypothesis test of exact accuracy (success rate) over varying sample sizes, i.e., from 100 to 1000 samples. The solid lines indicate the exact accuracy, while the shaded areas indicate the standard error. The results show that for all models, (1) the accuracy stabilizes after $500$ samples, and (2) the standard error drops significantly as sample sizes increase from $100$ to $1000$ samples. This demonstrates (1) the necessity of using larger sample size and (2) the sufficiency of using a sample size of $1000$, to achieve reliable evaluation (see~\appref{app:results} for details and more experiments on statistical significance). 

\section{Conclusion}\label{sec:conclusion}

We propose MMNeedle, a benchmark to evaluate MLLMs' long-context capabilities. MMNeedle includes a comprehensive dataset and establishes diverse settings as well as a systematic set of coarse-to-fine evaluation metrics. 
We reveal that while API-based models, such as GPT-4o, outperform open-source models in long-context scenarios, they still struggle with hallucination issues in negative samples and challenges in large stitching size/multi-needle retrieval. A limitation of our MMNeedle evaluation is the assumption that the MLLM takes both images and texts as inputs and supports multiple-image inputs. However, we argue that these are necessary requirements for an ideal MLLM.

\section{Ethical Considerations} 
Our MMNeedle dataset, created from MS COCO images, adheres to ethical guidelines and ensures that the usage of images is respectful and does not infringe on personal privacy. We ensure that MMNeedle dataset does not contain any personally identifiable information or offensive content. We bear all responsibility in case of violation of rights and confirm that we use the CC BY 4.0 data license.

{Despite these precautions, there remains a risk that the benchmark's capabilities could be misused, particularly in scenarios where models are pushed to handle extensive visual contexts that may lead to unintended inferences or biases. Additionally, the risk of hallucination in negative samples, where the model incorrectly identifies a nonexistent target, highlights the importance of responsible use and the need for thorough evaluation before deploying these models in high-stakes applications.}

\section{Limitations}
{Our MMNeedle Benchmark assumes that the evaluated MLLM can understand and follow both visual and textual instructions, and that the model can process multiple images as input in a single query. While this is not general, we note that these assumptions (and capabilities) are necessary for modern, state-of-the-art MLLMs.} {Adding textual or visual index labels next to each image or sub-image could potentially enhance the performance of models. However, we leave this exploration for future work for the following reasons: (1) Our MMNeedle's goal is to measure MLLM's long-context capability on natural images. Accuracy of predicting sub-image indices serves as one way of measuring such capabilitiy, but the accuracy itself is not the final goal. (2) This approach alters the original image content. MMNeedle is also limited by the supported number $M$ and stitching size $N$ of image inputs in MLLMs. However, our framework can seamlessly accommodate larger  $M$ and $N$ once open-source and API models (e.g., GPT-4o) begin to support them. }

\section{{Acknowledgements}}
We sincerely appreciate the generous support from the Microsoft Research AI \& Society Fellowship, NSF Grant IIS-2127918, NSF CAREER Award IIS-2340125, NIH Grant 1R01CA297832, and the Amazon Faculty Research Award. This research is also supported by NSF National Artificial Intelligence Research Resource (NAIRR) Pilot and the Frontera supercomputer, funded by the National Science Foundation (award NSF-OAC 1818253) and hosted at the Texas Advanced Computing Center (TACC) at The University of Texas at Austin. Finally, we extend our gratitude to the Center for AI Safety (CAIS) for providing the essential computing resources that made this work possible.

\bibliography{custom}
\appendix


\section{Details of the MMNeedle Dataset}\label{app:dataset}
\begin{figure*}[h]
        \centering
    \includegraphics[width = 0.99\textwidth]
        {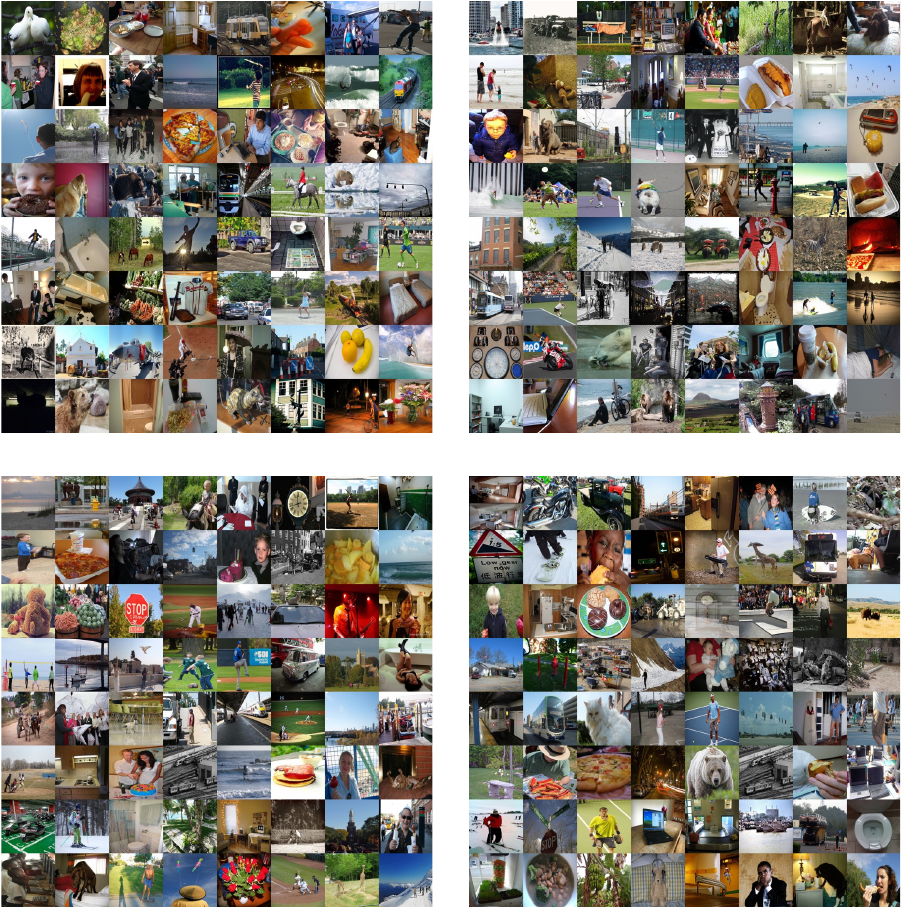}
        \caption[width = 0.8\linewidth]{Random samples of $8 \times 8$ stitched images in the MMNeedle dataset.}
        \label{fig:random_samples}
        \vskip -0.2cm
\end{figure*}

We include all the images, captions, prompts, and needle-haystack pairs of our MMNeedle Dataset at~\href{https://github.com/Wang-ML-Lab/multimodal-needle-in-a-haystack}{https://github.com/Wang-ML-Lab/multimodal-needle-in-a-haystack}. 

\textbf{{Limits on the Image Numbers.}}
We set the maximum number of complete images to $M=10$ because this is the largest number of input images that GPT-4V/4o can support. Note that our framework can easily handle larger $N$ and $M$ once open-source and API models (e.g., GPT-4o) start to support them. 
Table~\ref{tab:maximum_images} below summarizes each API-based model's limit for the number of images. 


\begin{table}[h!]
    \centering
    \caption{Maximum number of images per request. "*" indicates that the OpenAI GPT-4V/4o API also supports a maximum of $10$ images with high quality. Other numbers are \emph{hard} limits.}
    \begin{tabular}{ll}
        \toprule
        \textbf{API-Based Model} & \textbf{Limit} \\
        \midrule
        Azure GPT-4V/4o & 10 \\
        OpenAI GPT-4V/4o & 10$^*$ \\
        Claude 3 Opus & 20 \\
        Gemini 1.0 Pro & 16 \\
        \bottomrule
    \end{tabular}
    \label{tab:maximum_images}
    \vskip -0.3 cm
\end{table}
It is worth noting that:
\begin{itemize}
\item Azure OpenAI API only supports $10$ images for GPT-4V/4o. For example, an Azure document states that ``When uploading images, there is a limit of 10 images per chat request.'' Another Azure document states that ``GPT-4o max images per request'' is $10$. 
\item Regular OpenAI API also supports a maximum of 10 images with high quality. Specifically, an OpenAI document states that ``the token cost of a given image is determined by two factors: its size, and the detail option on each image\_url block''. Therefore, to ensure sufficient quality/resolution of image inputs, we cannot upload more than $10$ images to GPT-4V/4o in the MMNeedle benchmark. 
\item Other models also have a limit on the number of input images (e.g., $20$ for Claude and $16$ for Gemini). Specifically, the Claude 3 Opus document states that ``You can include multiple images in a single request (up to 5 for claude.ai and 20 for API requests)'', and the Gemini 1.0 Pro Vision supports up to ``16 images'' as ``Maximum number of images per request''.
\end{itemize}

Therefore, to ensure a fair comparison, we conducted all multi-image experiments on the $M=10$ images setting.

\textbf{{Purpose of Image Stitching.}}
The reason we introduce stitching with $N\times N > 1\times 1$ is as follows:
\begin{itemize}
  \item API-based models, such as GPT-4V/4o, can support at most $10$ images as inputs, which is \emph{surprisingly small}. To further evaluate long contexts with more images, we decided to introduce image stitching. As a result, when $M=10, N\times N = 8\times 8$, there are equivalently $640$ sub-images in the context, which is sufficiently large compared to the API limits of a few images.
 \item Image stitching enables us to conduct additional evaluation on MLLMs' capability in localization and retrieval of sub-images within the complete input images, which is another important aspect of long-context problems.
\end{itemize}
\textbf{Resolution of Sub-Images.}
As discussed in~\secref{sec:construct_long_context} of the main paper, we find that humans and LLMs cannot effectively recognize MS COCO images with a resolution lower than $256$. \figref{fig:random_samples} shows 4 random samples with $8\times 8$ stitching from our MMNeedle dataset. As demonstrated in these images, our $256\times 256$ resolution ensures a reasonable balance of input tokens and image quality. {Consequently, for a stitch size of $N \times N$, the overall resolution becomes $256N \times 256N$, resulting in a longer input context length that scales linearly with the stitch size $N$. This approach ensures that we do not downsample the sub-images in the stitched image, while still maintaining high image quality for the model's comprehension.} {The Azure OpenAI document} states that: ``If an image is ambiguous or unclear, the model will do its best to interpret it. However, the results might be less accurate. A good rule of thumb is that if an average human can't see the info in an image at the resolutions used in low/high res mode, then the model can't either.'' {The Anthropic document} also states that ``Ensure your images are clear and not too blurry or pixelated. Claude may struggle to accurately interpret unclear or low-quality images.'' {Indeed, our stitched images demonstrate sufficiently high resolution to be recognized by both humans and MLLMs, and there is very little content loss or noise introduced.}


\textbf{Data Source.} The asset we use in our paper, i,e, MS COCO 2014 dataset, is licensed under a Creative Commons Attribution 4.0 License. This license permits the copying, redistribution, remixing, transforming, and building upon the material for any purpose, including commercial use, provided appropriate credit is given, and any changes made are indicated. As a user of the MS COCO dataset, we acknowledge and comply with the requirements of the CC BY 4.0 license.




\textbf{Evaluation Metrics for Multiple Needles.} As mentioned in~\secref{sec:metrics} of the main paper, we use similar metrics for the multi-needle setting:
\begin{itemize}
\item \textbf{Existence Accuracy} is the proportion of samples in which the model correctly predicts whether \emph{any needle exists}, i.e., at least one target caption matches a sub-image in the input image sequence.
\item \textbf{Index Accuracy} is the proportion of samples where the model correctly predicts the index $m\in \{1,...,M\}$ of the stitched image containing the needle \emph{for all the needles}.
\item \textbf{Exact Accuracy} is the proportion of samples where the model correctly predicts the needle sub-image’s location, i.e., index $m$, row $r$ and column $c$ \emph{for all the needles}.
\end{itemize}

In this paper, we evaluate MLLMs with the number of needles $K\in \{1, 2, 5\}$. Our primary evaluation involves testing on the first $1000$ positive and the first $1000$ negative samples in our dataset using a single needle. As complementary experiments, we also test multi-needle settings with $2$ and $5$ needles on the first $100$ positive and the first $100$ negative samples in our dataset, respectively. Due to time and rate limits, as well as the high cost of testing API models, we are able to test $2000$ samples for each single-needle setting and $200$ samples for each multi-needle setting. However, our test easily scale to more samples, such as other samples in our $10{,}000$-sample dataset. We also show that the accuracy stabilizes when the test number reaches $1000$ in~\secref{sec:stat} of the main paper and~\appref{app:stat}. 




\textbf{Prompt Design}\label{sec:instruction}
For single-needle evaluation, we use the following prompt for the evaluated LLM: 
\begin{tcolorbox}
{\small\begin{verbatim}
Input = [Images] + Instructions + "\n" 
        + "Caption: " + Caption
\end{verbatim}
}
\end{tcolorbox}
where the \emph{instructions} to MLLMs is as follows:

\begin{tcolorbox}
Given $M$ images indexed from $1$ to $M$, 
each divided into $N\times N$ sub-images, 
identify the sub-image that best matches 
the provided caption. Respond with ``index, row, column'' and nothing else. 
For example, ``1, 2, 3'' indicates the sub-image in the 
first image, second row, and third column. 
If no match is found, respond only with ``-1''.
\end{tcolorbox}

We use a similar prompt for the multi-needle setting. 
Specifically, for $K$-needle ($K>1$) evaluation, we use the following prompt for the evaluated MLLM:
\begin{tcolorbox}
\tiny\begin{verbatim}
Input = [Images] + Instructions + "\n" 
    + "Caption 1: " + Caption_1 + "\n"+ "Caption 2: " + Caption_2 
    + "\n" + ... + "Caption K: " + Caption_K,
\end{verbatim}
\end{tcolorbox}
where the \emph{instructions} to MLLMs is as follows:

\begin{tcolorbox}
Given $M$ images indexed from $1$ to $M$, each divided into $N\times N$ sub-images, 
identify the sub-images that best match the provided K captions. 
Respond in the format: ``index\_1, row\_1, column\_1; ...; index\_K, row\_K, column\_K.'' Only provide this information. 
For example, ``1, 2, 3'' indicates the sub-image in the first image, second row, and third column. 
If no sub-image matches a caption, respond with ``-1'' for that caption.
\end{tcolorbox}

Note that for both single-needle and multi-needle settings, when $M=1$ or $N=1$, we remove the ``s'' in ``images'' or ``sub-images'' in our prompt for coherent description, respectively.

\section{Details of Evaluation Process}
\label{app:protocol}
\textbf{Automated Evaluation Protocol.} 
As discussed in~\secref{sec:metrics} of the main paper, we design an automated evaluation protocol for the three defined metrics as follows:
\begin{itemize}
    \item \textbf{Ground Truth Format.} For each caption in a test sample, (1) if it is positive, i.e., the needle sub-image is in the context, the ground-truth output is {``$m, r, c$''} that describes the location of the needle, where $m$ is the image index ($m \in \{1,...,M\}$), and $r, c$ are the row and column of the sub-image (needle) in image $m$, respectively ($r, c \in \{1,...,N\}$); (2) if it is negative, meaning no needle sub-image is in the context, the ground truth output is {``-1''}, indicating the needle does not exist.  
    For multi-needle settings, the ground truth is a concatenation of the ground-truth answer for each needle in the order of input captions, separated by {``;''}. For example, for a 2-needle test with $M=10$ and $N=8$, a positive answer can be ``1, 2, 8; 10, 3, 5'' and a negative answer should be ``-1; -1''. 
    \item \textbf{Existence Accuracy} is measured by whether the MLLM outputs ``-1'' 
    (in multi-needle settings, we match ``-1'' for all the needles, separated by ``;'', or alternatively just one ``-1''). Specifically, for positive samples (targets exist), the existence accuracy is the proportion of samples where the MLLM does not predict ``-1'', and for negative samples (targets do not exist), the existence accuracy is the proportion of of samples where the MLLM predicts ``-1''.
    \item \textbf{Index Accuracy} is measured by whether the image index $\hat{m}$ predicted by the MLLM matches the ground truth $m$. For multi-needle settings, predictions are considered correct only if the MLLM predicts the correct $m$ for all needles. Note that even for the $M=1$ settings, the index accuracy may not be perfect ($100\%$), because the model can fail to output the correct image index ``1''. Therefore, we also evaluate the index accuracy of different models in the $M=1$ settings.
    \item \textbf{Exact Accuracy} is measured by whether the tuple $(\hat{m}, \hat{r}, \hat{c})$ predicted by the MLLM matches the ground truth $(m, r, c)$. For multi-needle settings, predictions are considered correct only if the MLLM predicts the correct $(m, r, c)$ for all needles. 
    \item \textbf{(Multi-Needle) Individual Accuracy} is measured by whether the tuple $(\hat{m}, \hat{r}, \hat{c})$ predicted by the MLLM matches the ground truth $(m, r, c)$ in multi-needle samples, where predictions are considered correct only if the MLLM predicts the correct $(m, r, c)$ for each individual needle. 
    
\end{itemize}

This automated evaluation protocol can be seamlessly integrated with prompt design, where our prompts ask the MLLM to output in the format of the ground truth. As discussed in~\secref{sec:overview} of the main paper, the model can successfully produce a correct answer only if it understands our instructions, recognizes where there are needles in the haystack that match the given text query (target captions), and outputs in the correct format. Otherwise, the MLLM may produce answers with incorrect formats or meanings, resulting in failed cases.

Our multimodal evaluation benefits from canonical ground-truth answers and is therefore not affected by the similarity of the needles to test and data points in the training set in terms of output tokens.
 \begin{itemize}
     \item[(1)] Compared to other open-ended evaluations, since we ask the MLLMs to output the locations of the target sub-images, the model has no back-doors to output a ``seemingly'' correct answer as in other open-ended generation. These back-doors include learning the next token distribution from the training set and responding with the contents of other images.
     \item[(2)] Compared to multiple-choice questions, the chance that the model outputs coincidentally match the correct answer is also much lower. For example, the accuracy of a random guess in 4-choice problems is always 25\%, while even in our easiest settings (1 image, $2 \times 2$ stitching; 10 images, $1 \times 1$ stitching), the accuracy is 25\% and 10\%, respectively.
 \end{itemize}

\textbf{Post-Processing.} In Table 3 of the main paper, IDEFICS2-8B $M=1,N=4$ results on negative samples are as low as $20.20 \%$ due to its failure to follow instructions on the output format, particularly affected by the ``Answer: '' prefix in responses. Therefore, we include additional parsing for this case, resulting in an accuracy of $55.70 \%$ in the same setting. Specifically, we use additional filtering of the prefix ``Answer:'' for IDEFICS2-8B in $M=1,N=4$ negative samples. 

\section{Implementation Details}\label{app:implement}
All the code, data, and instructions required to reproduce the main experimental results are provided in the supplementary materials (``Software'' and ``Data'').

\textbf{Compute and Resources.}
For the API-based models, we used the corresponding API credits to conduct our experiments: Anthropic API for Claude 3 Opus, Google Cloud API for Gemini Pro 1.0 and Gemini Pro 1.5, and Azure OpenAI API service for GPT-4V and GPT-4o. For the open-source models, we used 2 Nvidia A100 GPUs for our evaluation. Each model required a few hours to a few days to complete the evaluation, depending on the API rate limit or GPU memory limit.

\textbf{Model Details.} As discussed in~\secref{sec:exp_model} of the main paper, we conduct MMNeedle evaluation for both API-based models and open-source models:
\begin{itemize}
    \item \textbf{API-based models} are state-of-the-art multimodal LLMs with API calling access:
    \begin{itemize}
        \item \textbf{Claude 3 Opus}~\cite{anthropic2023claude3} is the \emph{strongest} MLLM developed by Anthropic. We use the model version \emph{claude-3-opus-20240229}. 
        \item \textbf{Gemini Pro 1.0}~\cite{team2023gemini} is an advanced version of Google Gemini, offering enhanced performance in multimodal tasks. We use the model version \emph{gemini-1.0-pro-vision-latest}.
        \item \textbf{Gemini Pro 1.5}~\cite{reid2024gemini} is built upon Gemini Pro 1.0 with further optimizations in multimodal capability, serving as the \emph{strongest} model version of Google Gemini. We use the model version \emph{gemini-1.5-pro-latest}.
        \item \textbf{GPT-4V}~\cite{achiam2023gpt4} is an extension of OpenAI's GPT-4, equipped with vision capabilities for multimodal tasks. We use Azure OpenAI API with the model version \emph{2024-03-01-preview}.
        \item \textbf{GPT-4o}~\cite{openai2024gpt4o} is the \emph{latest and strongest} variant of OpenAI's GPT-4. We use Azure OpenAI API with the model version \emph{2024-05-01-preview}.  
    \end{itemize}
    \item \textbf{Open-source models} are state-of-the-art methods with open access to their weights:
    \begin{itemize}
        \item \textbf{CogVLM}~\cite{wang2023cogvlm} is a state-of-the-art MLLM for \emph{single-image} inputs. We evaluate CogVLM-17B-base and CogVLM2-Llama-3 (the \emph{latest and strongest} version).
        \item \textbf{Fuyu-8B}~\cite{fuyu-8b} is a state-of-the-art, 8-billion-parameter model that excels in multimodal tasks compared to other models of similar size.
        \item \textbf{mPLUG-Owl-v2}~\cite{ye2023mplug} is an updated version of mPLUG-Owl and also a state-of-the-art MLLM.
        \item \textbf{InstructBLIP}~\cite{dai2024instructblip} is another state-of-the-art MLLM for \emph{single-image} inputs. We evaluate InstructBLIP-Vicuna-13B and {InstructBLIP-Flan-T5-XXL}, which are its two strongest variants.
        \item \textbf{IDEFICS2}~\cite{laurenccon2024idefics2} is the latest version of IDEFICS and also a state-of-the-art MLLM.
        \item \textbf{LLaVA-Llama-3}~\cite{li2024llavanext-strong} is the \emph{latest and strongest} version of LLaVA~\cite{liu2024llava} and also a state-of-the-art MLLM.

    \end{itemize}
\end{itemize}

\textbf{Samples Skipped by API-based Models.} Due to the built-in filters for the API-based models, they may refuse to answer questions for a small number of samples in our dataset. However, the number of refused questions is limited to dozens out of $2{,}000$ samples in each setting. Therefore, excluding these vacant samples in the results does not affect any of our conclusions. See the statistical significance discussion in~\appref{app:stat}, as well as~\secref{sec:stat} of the main paper.

\section{More Experimental Results}\label{app:results}
\begin{figure}[t]
        \centering
    \includegraphics[width = 0.48\textwidth]
        {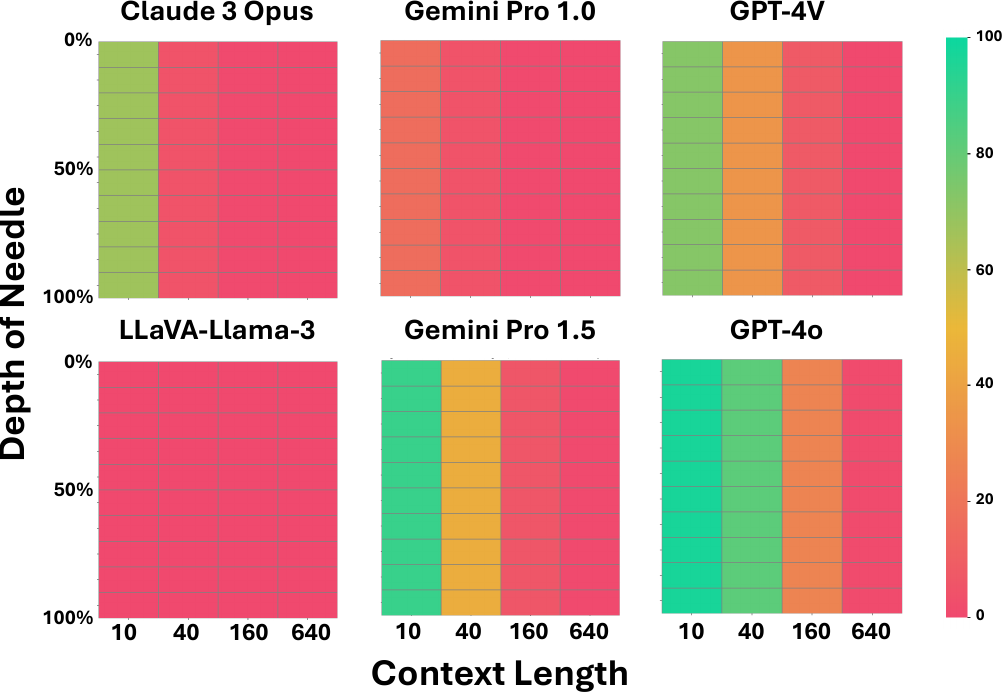}
        \vskip -0.25cm
        \caption[width = 0.8\linewidth]{Accuracy~(\%) under different needle depths and context lengths on $M=10$ samples. A \emph{\red{redder}} cell indicates lower accuracy, while a \emph{\green{greener}} cell indicates higher accuracy.}
        \label{fig:needle_depth}
        \vskip -0.3cm
\end{figure}

\begin{table*}[t]
\caption{Exact Accuracy $\pm$ Standard Error (\%) of GPT-4V for the 1-needle samples with different instruction structures. We mark the best results with \textbf{bold face}.}\label{tab:gpt4_reorder_needle1}
\vskip  0.1 cm
\centering
\resizebox{0.99\textwidth}{!}{%
\begin{tabular}{lccccccc}
\toprule
Stitching & \multicolumn{1}{c}{$1\times1$} & \multicolumn{2}{c}{$2\times2$} & \multicolumn{2}{c}{$4\times4$} & \multicolumn{2}{c}{$8\times8$}  \\
\cmidrule(r){1-1} \cmidrule(r){2-2} \cmidrule(r){3-4} \cmidrule(r){5-6} \cmidrule(r){7-8} 
Instructions & 10 imgs & 1 img& 10 imgs& 1 img& 10 imgs &1 img & 10 imgs \\ 
\midrule
Prompt + Caption & \textbf{74.49}$\pm$4.36& \textbf{85.71}$\pm$3.50 & 30.21$\pm$4.59 & 45.00$\pm$4.97 & \textbf{8.16}$\pm$2.74& 8.00$\pm$2.71 & \textbf{0.00}$\pm$0.00\\
Caption + Prompt& \textbf{74.49}$\pm$4.36& 80.61$\pm$3.95&\textbf{33.33}$\pm$4.71 &\textbf{49.00}$\pm$5.00 & 5.10$\pm$2.20 &\textbf{9.00}$\pm$2.86 & \textbf{0.00}$\pm$0.00\\
\bottomrule
\end{tabular}%
}
\vskip -0.5cm
\end{table*}
\textbf{Effect of Needle Depth.}
We investigated the effect of needle depth on the accuracy of MLLMs. Specifically, we tested different needle depths ranging from $1$ to $10$ for $M=10$ images in a single-needle setting. We calculated the accuracy for each depth, analyzing how well the models could identify the correct needle image across various depths. \figref{fig:needle_depth} shows the accuracy of models on different needle depths and context lengths. The results show that for all models, accuracy drops significantly with increasing context lengths, while the accuracy of different needle depths shows little variation for the same model and context length.

\begin{figure}[t]
        \centering
    \includegraphics[width = 0.48\textwidth]
        {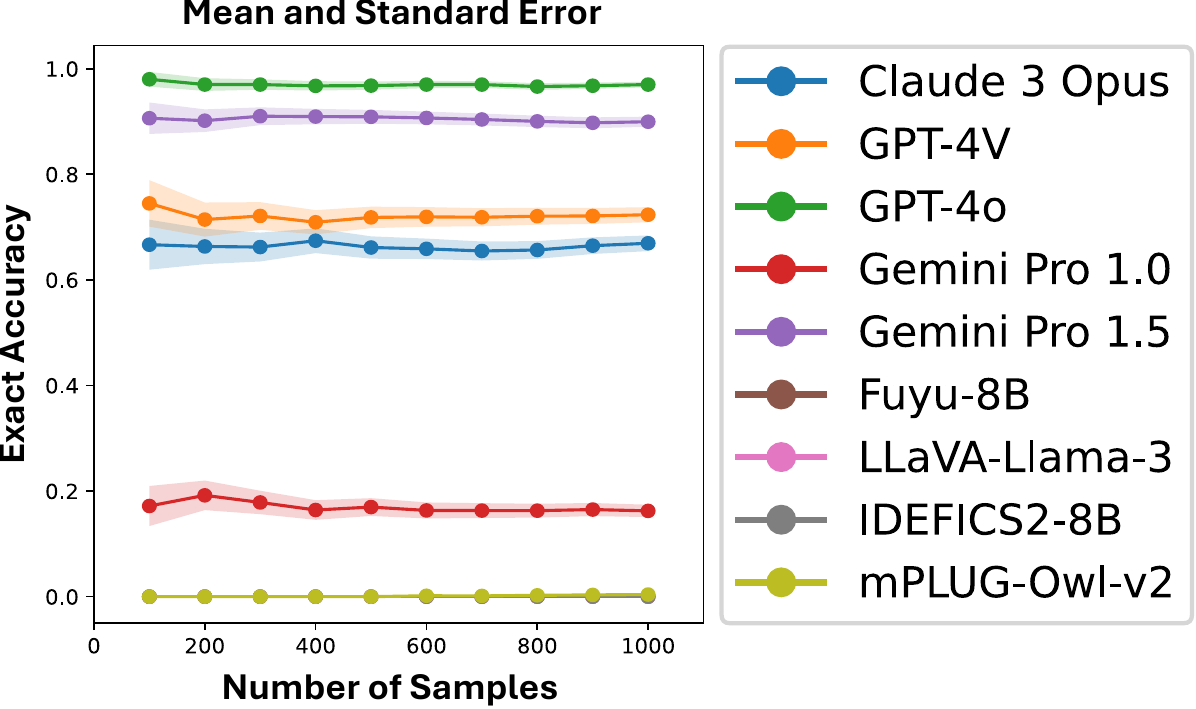}
        \vskip -0.25cm
        \caption[width = 0.8\linewidth]{Exact Accuracy and Standard Error of Different Models on $M=10, N=1$ Samples. The accuracies of all open-source models on these samples are very close to ${0\%}$.}
        \label{fig:10imag_statistics}
        \vskip -0.3cm
\end{figure}

\textbf{Statistical Significance.}\label{app:stat}
To ensure the robustness of our evaluation, we conducted hypothesis tests for the exact accuracy (mean of binary value for each sample) of different models under the binomial distribution $\text{Binomial}(1,p)$, where $p$ is the probability of success on an individual trial. The standard error (SE) of this test is calculated as follows:
\begin{align}
    \text{SE} = \sqrt{\frac{p(1 - p)}{s}},
\end{align}
where $s$ is the number of trials (samples). \figref{fig:10imag_statistics} shows the mean and standard error of exact accuracy for different models in the $M=1, N=10$ setting. Note that InstructBLIP and CogVLM models do not support multi-image inputs; therefore we exclude them in the figure. The results indicate that the accuracy stabilizes after approximately 500 samples, and the standard error decreases significantly as the sample size increases from 100 to 1000. This highlights the importance of utilizing larger sample sizes to ensure reliable evaluation results, as discussed in~\secref{sec:stat} of the main paper.



\textbf{Effect of the Instruction Order.}
\tabref{tab:gpt4_reorder_needle1} shows the exact accuracy of the GPT-4V model in each different $M,N$ setting on $100$ random positive samples. ``Prompt+Caption (default)'' means our prompt is followed by a caption in the instructions, and ``Caption+Prompt (alternative)'' means a caption is followed by our prompt in the instructions. The results indicate that these two different ordered instructions are not statistically significantly better than each other for any setting.

\begin{table*}[t]
\caption{Accuracy (\%) in the three metrics for the 5-needle, $M=1$ samples. We mark the best results with \textbf{bold face}. Note that the existence accuracy is measured by whether the model outputs ``-1'' for all the needles. The index accuracy is not always $100$ \% because the model can fail to output the correct image index ``1''.}\label{tab:img1_needle5}
\vskip  0.2 cm
\centering
\resizebox{0.99\textwidth}{!}{%
\begin{tabular}{lccccccccccccccc}
\toprule
Stitching & \multicolumn{3}{c}{$2\times2$} & \multicolumn{3}{c}{$4\times4$} & \multicolumn{3}{c}{$8\times8$}  \\
\cmidrule(r){1-1} \cmidrule(r){2-4} \cmidrule(r){5-7} \cmidrule(r){8-10} 
Metrics & \emph{Existence} & \emph{Index} & \emph{Exact} & \emph{Existence} & \emph{Index} & \emph{Exact} & \emph{Existence} & \emph{Index} & \emph{Exact}\\ 
\midrule
{API-based models}\\
\midrule
Claude 3 Opus & \bf{100.00}&22.00&2.00 & \bf{100.00}&37.00&0.00 & \bf{100.00}&29.00&0.00\\
Gemini Pro 1.0 &\bf{100.00}&32.00&1.00&\bf{100.00}&6.00&0.00&\bf{100.00}&0.00&0.00\\
Gemini Pro 1.5& \bf{100.00}&\bf{91.00}&24.00 &\bf{100.00}&\bf{91.00}&1.00& \bf{100.00}&\bf{81.00}&0.00\\
GPT-4V & \bf{100.00}&55.91&\bf{34.41}&\bf{100.00}&68.37&\bf{8.16}&\bf{100.00}&61.62&0.00\\
GPT-4o& \bf{100.00}&28.00&24.00&\bf{100.00}&24.00&6.00&\bf{100.00}&22.00&0.00\\
\midrule
Open-source models\\
\midrule
CogVLM-17B& \bf{100.00}& 0.00 &0.00 & \bf{100.00}& 0.00 &0.00& \bf{100.00}& 0.00 &0.00\\
CogVLM2-LLaMA-3& \bf{100.00}& 0.00 &0.00 & \bf{100.00}& 1.00 &0.00& \bf{100.00}& 0.00 &0.00\\
Fuyu-8B & \bf{100.00}& 0.00 &0.00 & \bf{100.00}& 0.00 &0.00& \bf{100.00}& 0.00 &0.00\\
mPLUG-Owl-v2 & 98.00&0.00&0.00&98.00&2.00&0.00&98.00&0.00&0.00\\
InstructBLIP-Vicuna-13B & \bf{100.00}& 0.00 &0.00 & \bf{100.00}& 0.00 &0.00& \bf{100.00}& 0.00 &0.00\\
InstructBLIP-Flan-T5-XXL& \bf{100.00}& 0.00 &0.00 & \bf{100.00}& 0.00 &0.00& \bf{100.00}& 0.00 &0.00\\
IDEFICS2-8B & \bf{100.00}& 0.00 &0.00 & \bf{100.00}& 0.00 &0.00& \bf{100.00}& 0.00 &0.00\\
LLaVA-LLaMA-3  & \bf{100.00}& 3.00 &0.00 & \bf{100.00}& 2.00 &0.00& \bf{100.00}& 2.00 &0.00\\

\bottomrule
\end{tabular}%
}
\end{table*}

\begin{table*}[!t]
\vskip -0.3cm
\caption{Accuracy (\%) in the three metrics for the 2-needle, $M=10$ samples. We mark the best results with \textbf{bold face}. Note that the existence accuracy is measured by whether the model outputs ``-1'' for all the needles.}\label{tab:img10_needle2}
\vskip  0.2 cm
\centering
\resizebox{0.99\textwidth}{!}{%
\begin{tabular}{lcccccccccccccccccc}
\toprule
Stitching & \multicolumn{3}{c}{$1\times1$} & \multicolumn{3}{c}{$2\times2$} & \multicolumn{3}{c}{$4\times4$} & \multicolumn{3}{c}{$8\times8$}  \\
\cmidrule(r){1-1} \cmidrule(r){2-4} \cmidrule(r){5-7} \cmidrule(r){8-10} \cmidrule(r){11-13} 
Metrics & \emph{Existence} & \emph{Index} & \emph{Exact} & \emph{Existence} & \emph{Index} & \emph{Exact} & \emph{Existence} & \emph{Index} & \emph{Exact}& \emph{Existence} & \emph{Index} & \emph{Exact}\\ 
\midrule
{API-based models}\\
\midrule
Claude 3 Opus & \bf{100.00}&46.00&46.00&\bf{100.00}&1.12&0.00&98.00&0.00&0.00&96.91&1.03&0.00\\
Gemini Pro 1.0 & 92.93&3.03&0.00&98.00&1.00&0.00&\bf{100.00}&0.00&0.00&99.00&0.00&0.00\\
Gemini Pro 1.5& \bf{100.00}&{86.73}&{85.71}&\bf{100.00}&{34.00}&{25.00}&\bf{100.00}&2.08&0.00&85.86&0.00&0.00\\
GPT-4V &\bf{100.00}&52.17&48.91&\bf{100.00}&25.58&6.98&\bf{100.00}&{3.45}&0.00&\bf{100.00}&{1.19}&0.00\\
GPT-4o& \bf{100.00}&\bf{88.00}&\bf{88.00}&\bf{100.00}&\bf{71.00}&\bf{53.00}&\bf{100.00}&\bf{13.00}&\bf{5.00}&\bf{100.00}&\bf{3.00}&0.00\\
\midrule
Open-source models\\
\midrule
Fuyu-8B & \bf{100.00}& 0.00 &0.00 & \bf{100.00}& 0.00 &0.00& \bf{100.00}& 0.00 &0.00& \bf{100.00}& 0.00 &0.00\\
mPLUG-Owl-v2 & 66.00&0.00&0.00&90.00&0.00&0.00&97.00&0.00&0.00&96.00&0.00&0.00\\
IDEFICS2-8B &59.00& 0.00 &0.00  & 94.00& 0.00 &0.00 & \bf{100.00}& 0.00 &0.00& 99.00& 0.00 &0.00\\
LLaVA-LLaMA-3&  \bf{100.00}& 0.00 &0.00  & \bf{100.00}& 0.00 &0.00 & \bf{100.00}& 0.00 &0.00& \bf{100.00}& 0.00 &0.00\\

\bottomrule
\end{tabular}%
}
\end{table*}

\begin{table*}[!t]
\caption{Accuracy (\%) in terms of the three metrics for the 5-needle, $M=10$ samples. We mark the best results with \textbf{bold face}. Note that the existence accuracy is measured by whether the model outputs ``-1'' for all the needles.}\label{tab:img10_needle5}
\vskip  0.2 cm
\centering
\resizebox{0.99\textwidth}{!}{%
\begin{tabular}{lcccccccccccccccccc}
\toprule
Stitching & \multicolumn{3}{c}{$1\times1$} & \multicolumn{3}{c}{$2\times2$} & \multicolumn{3}{c}{$4\times4$} & \multicolumn{3}{c}{$8\times8$}  \\
\cmidrule(r){1-1} \cmidrule(r){2-4} \cmidrule(r){5-7} \cmidrule(r){8-10} \cmidrule(r){11-13} 
Metrics & \emph{Existence} & \emph{Index} & \emph{Exact} & \emph{Existence} & \emph{Index} & \emph{Exact} & \emph{Existence} & \emph{Index} & \emph{Exact}& \emph{Existence} & \emph{Index} & \emph{Exact}\\ 
\midrule
{API-based models}\\
\midrule
Claude 3 Opus & \bf{100.00}&{32.32}&{32.32}&\bf{100.00}&0.00&0.00&\bf{100.00}&0.00&0.00&\bf{100.00}&0.00&0.00
\\
Gemini Pro 1.0 & \bf{100.00}&0.00&0.00&\bf{100.00}&0.00&0.00&\bf{100.00}&0.00&0.00&\bf{100.00}&0.00&0.00\\
Gemini Pro 1.5& \bf{100.00}& \bf{82.83}&13.13&\bf{100.00}&{7.00}&0.00&\bf{100.00}&0.00&0.00&\bf{100.00}&0.00&0.00\\
GPT-4V &\bf{100.00}&28.12&25.00&\bf{100.00}&1.14&0.00&\bf{100.00}&0.00&0.00&\bf{100.00}&0.00&0.00\\
GPT-4o& \bf{100.00}&73.00&\bf{69.00}&\bf{100.00}&\bf{37.00}&\bf{8.00}&\bf{100.00}&0.00&0.00&\bf{100.00}&0.00&0.00\\
\midrule
Open-source models\\
\midrule
Fuyu-8B & \bf{100.00}& 0.00 &0.00  & \bf{100.00}& 0.00 &0.00 & \bf{100.00}& 0.00 &0.00& \bf{100.00}& 0.00 &0.00\\
mPLUG-Owl-v2  & 82.00&0.00&0.00&93.00&0.00&0.00&97.00&0.00&0.00&\bf{100.00}&0.00&0.00\\

IDEFICS2-8B& 69.00& 0.00 &0.00  & 91.00& 0.00 &0.00 & 98.00& 0.00 &0.00& 99.00& 0.00 &0.00\\
LLaVA-LLaMA-3& \bf{100.00}& 0.00 &0.00  & \bf{100.00}& 0.00 &0.00 & \bf{100.00}& 0.00 &0.00& \bf{100.00}& 0.00 &0.00\\

\bottomrule
\end{tabular}%
}
\end{table*}

\begin{table*}[!t]
\vskip 0cm
\caption{Existence Accuracy (\%) for the 2-needle negative samples (the ground truth is ``-1; -1''). We mark the best results with \textbf{bold face}. Note that the existence accuracy is measured by whether the model outputs ``-1'' for all the needles. ``-'' means that the models do not support multi-image inputs.}\label{tab:neg_needle2}
\vskip  0.2 cm
\centering
\resizebox{0.79\textwidth}{!}{%
\begin{tabular}{lccccccc}
\toprule
Stitching & \multicolumn{1}{c}{$1\times1$} & \multicolumn{2}{c}{$2\times2$} & \multicolumn{2}{c}{$4\times4$} & \multicolumn{2}{c}{$8\times8$}  \\
\cmidrule(r){1-1} \cmidrule(r){2-2} \cmidrule(r){3-4} \cmidrule(r){5-6} \cmidrule(r){7-8} 
Context & 10 imgs & 1 img& 10 imgs& 1 img& 10 imgs &1 img & 10 imgs \\ 
\midrule
{API-based models}\\
\midrule
Claude 3 Opus & 45.00&14.00&0.00&5.00&1.00&4.00&8.33\\
Gemini Pro 1.0 & 54.64&\bf{85.86}&18.00&{50.00}&0.00&\bf{34.00}&0.00\\
Gemini Pro 1.5& {79.59}&71.00&\bf{31.00}&{50.00}&\bf{7.37}&22.00&\bf{17.00}\\
GPT-4V & 74.75& 77.00&13.40 & 33.00&3.00 & 0.00&0.00\\
GPT-4o& \bf{80.00}&67.00&25.00&\bf{51.00}&3.00&2.00&0.00\\
\midrule 
Open-source models\\
\midrule
CogVLM-17B & -& 0.00& -& 0.00& -& 0.00& -\\
CogVLM2-LLaMA-3 & -& 0.00& -& 0.00& -& 0.00& -\\

Fuyu-8B& 0.00& 0.00& 0.00& 0.00& 0.00& 0.00& 0.00\\
mPLUG-Owl-v2  & 36.00&7.00&7.00&9.00&2.00&7.00&6.00\\
InstructBLIP-Vicuna-13B & -& 0.00& -& 0.00& -& 0.00& -\\
InstructBLIP-Flan-T5-XXL& -& 0.00& -& 1.00& -& 0.00& -\\

IDEFICS2-8B& 39.00& 0.00& 7.00& 0.00& 0.00& 0.00& 1.00\\
LLaVA-LLaMA-3& 0.00& 0.00& 0.00& 0.00& 0.00& 0.00& 0.00\\

\bottomrule
\end{tabular}%
}
\vskip 0cm
\end{table*}

\begin{table*}

\caption{Existence Accuracy (\%) for the 5-needle negative samples (the ground truth is ``-1; -1; -1; -1; -1''). We mark the best results with \textbf{bold face}. Note that the existence accuracy is measured by whether the model outputs ``-1'' for all the needles. ``-'' means that the models do not support multi-image inputs.}\label{tab:neg_needle5}
\vskip  0.2 cm
\centering
\resizebox{0.79\textwidth}{!}{%
\begin{tabular}{lccccccc}
\toprule
Stitching & \multicolumn{1}{c}{$1\times1$} & \multicolumn{2}{c}{$2\times2$} & \multicolumn{2}{c}{$4\times4$} & \multicolumn{2}{c}{$8\times8$}  \\
\cmidrule(r){1-1} \cmidrule(r){2-2} \cmidrule(r){3-4} \cmidrule(r){5-6} \cmidrule(r){7-8} 
Context & 10 imgs & 1 img& 10 imgs& 1 img& 10 imgs &1 img & 10 imgs \\ 
\midrule
{API-based models}\\
\midrule
Claude 3 Opus & 14.14&2.00&0.00&0.00&0.00&0.00&0.00\\
Gemini Pro 1.0 & 1.00&32.00&0.00&1.01&\bf{1.00}&0.00&0.00\\
Gemini Pro 1.5& 56.57&60.00&4.00&15.15&0.00&{1.00}&0.00\\
GPT-4V & \bf{73.63}&{65.96}&{8.99}&{17.00}&0.00&0.00&0.00\\
GPT-4o& 58.00& \bf{67.00}&\bf{25.00}&\bf{37.00}&0.00&\bf{2.00}&0.00\\
\midrule 
Open-source models\\
\midrule
CogVLM-17B & -& 0.00& -& 0.00& -& 0.00& -\\
CogVLM2-LLaMA-3 & -& 0.00& -& 0.00& -& 0.00& -\\
Fuyu-8B& 0.00& 0.00& 0.00& 0.00& 0.00& 0.00& 0.00\\
mPLUG-Owl-v2  & 40.00&5.00&2.00&5.00&3.00&2.00&3.00\\
InstructBLIP-Vicuna-13B & -& 0.00& -& 0.00& -& 0.00& -\\
InstructBLIP-Flan-T5-XXL& -& 0.00& -& 0.00& -& 0.00& -\\

IDEFICS2-8B& 29.00& 0.00& 12.00& 0.00& 0.00& 0.00& 0.00\\
LLaVA-LLaMA-3& 0.00& 0.00& 0.00& 0.00& 0.00& 0.00& 0.00\\

\bottomrule
\end{tabular}%
}
\vskip -0.2cm
\end{table*}

\textbf{Results on Multi-Needle Single-Image Samples.}
In additional to~\secref{sec:acc_results} of the main paper, \tabref{tab:img1_needle5} shows the accuracy on samples in the $M = 1, K=5$ setting, with three different stitching scenarios (i.e., $N\times N$ as $2\times 2$, $4\times 4$, and $8\times 8$). GPT-4V achieves the highest exact accuracy $34.41\%$ and $8.16\%$ for the $2\times 2$ and $4\times 4$ stitching, respectively, with accuracy dropping significantly to $0.00\%$ for the $8\times 8$ stitching. 
All open-source models show zero exact accuracy across all settings, falling behind in more needles ($K=5$) scenarios.

\textbf{Results on Multi-Needle Multi-Image Samples.}
\tabref{tab:img10_needle2} shows the accuracy on samples in the $M = 10, K=2$ setting, with four different stitching scenarios (i.e., $N\times N$ as $1\times1$, $2\times 2$, $4\times 4$, and $8\times 8$). GPT-4o achieves the highest exact accuracy of $88.00\%$ and $53.00\%$ for the $1\times 1$ and $2\times 2$ stitching, respectively, with accuracy dropping significantly to $5.00\%$ for the $4\times 4$ stitching.

\tabref{tab:img10_needle5} shows the accuracy on samples in the $M = 10, K=5$ setting, with four different stitching scenarios (i.e., $N\times N$ as $1\times1$, $2\times 2$, $4\times 4$, and $8\times 8$). GPT-4o achieves the highest exact accuracy of $69.00\%$ for the $1\times 1$ stitching, while its accuracy drops significantly to $8.00\%$ for the $2\times 2$ stitching.

All open-source models show zero exact accuracy across all settings, falling behind in more complex ($M=10$) scenarios. These results indicate the difficulty of our multi-needle multi-image evaluation.

\textbf{Results on Multi-Needle Negative Samples.}
\tabref{tab:neg_needle2} and \tabref{tab:neg_needle5} show the existence accuracy for negative samples in multi-needle settings ($K=2$ or $K=5$). In~\tabref{tab:neg_needle2}, representing the $K=2$ setting, Gemini Pro 1.5 achieves the highest existence accuracy in the $M=10$, $N\in\{2,4,8\}$ scenarios, indicating a low level of hallucination for long-context samples.
In contrast, in~\tabref{tab:neg_needle5}, representing the $K=5$ setting, GPT-4o achieves the best existence accuracy of $25.00\%$ and $37.00\%$ for $M=10,N=2$ and $M=1,N=4$ samples, respectively. 

The performance of open-source models fall behind in multi-needle negative samples, with mPLUG-Owl-v2 and IDEFICS2-8B performing better than others in both $K=2$ and $K=5$ settings.

\textbf{Results on Multi-Needle Individual Samples.}
\tabref{tab:indiv_img1_needle2}, \tabref{tab:indiv_img1_needle5}, \tabref{tab:indiv_img10_needle2}, and~\tabref{tab:indiv_img10_needle5} show the individual accuracy for multi-needle samples defined in~\appref{app:protocol}. Gemini Pro 1.5 achieves the highest exact accuracy for $N=2$ and $N=8$ samples in both~\tabref{tab:indiv_img1_needle2} and~ \tabref{tab:indiv_img1_needle5} (single-image inputs), while GPT-4o achieves the highest exact accuracy in both~\tabref{tab:indiv_img10_needle2} and~\tabref{tab:indiv_img10_needle5} (multi-image inputs).
\begin{table*}
\caption{Individual Accuracy (\%) in the three metrics for the 2-needle $M=1$ samples. We mark the best results with \textbf{bold face}. The index accuracy is not always $100$ \% because the model can fail to output the correct image index ``1''.}\label{tab:indiv_img1_needle2}
\vskip  0.1 cm
\centering
\resizebox{0.79\textwidth}{!}{%
\begin{tabular}{lcccccccccccc}
\toprule
Stitching & \multicolumn{2}{c}{$2\times2$} & \multicolumn{2}{c}{$4\times4$} & \multicolumn{2}{c}{$8\times8$}  \\
\cmidrule(r){1-1} \cmidrule(r){2-3} \cmidrule(r){4-5} \cmidrule(r){6-7} 
Metrics &   \emph{Index} & \emph{Exact}  & \emph{Index} & \emph{Exact}   & \emph{Index} & \emph{Exact}\\ 
\midrule
{API-based models}\\
\midrule
Claude 3 Opus & 82.01&49.74&55.62&10.00&49.67&2.61\\
Gemini Pro 1.0 &89.34&30.96&67.25&9.36&25.38&1.54\\
Gemini Pro 1.5& {97.47}&\bf{93.43}&92.00&42.50&89.00&\bf{26.00}\\
GPT-4V & 94.85&79.90&{97.00}&{56.00}&\bf{96.70}&5.49\\
GPT-4o& 96.28&86.17&96.81&\bf{74.47}&89.69&12.37\\
\midrule
Open-source models\\
\midrule
CogVLM-17B&  0.00 &0.00 &  0.00 &0.00& 0.00 &0.00\\
CogVLM2-LLaMA-3&  0.00 &0.00 &  0.00 &0.00& 0.00 &0.00\\
Fuyu-8B & 79.00& 0.00& 35.00&0.00&13.86&0.00\\
mPLUG-Owl-v2 & 41.77&1.27&14.75&0.00& 16.03&0.00\\
InstructBLIP-Vicuna-13B &  0.00 &0.00 & 0.00 &0.00& 4.00 &0.00\\
InstructBLIP-Flan-T5-XXL&  \bf{98.32} &24.37 &  \bf{100.00} &4.00& 75.00 &4.00\\
IDEFICS2-8B &  23.08 &0.96 &  84.40 &0.00& 14.00 &0.00\\
LLaVA-LLaMA-3  &  0.00 &0.00 &  13.00 &1.50& 25.50 &0.50\\

\bottomrule
\end{tabular}%
}
\vskip -0.3cm
\end{table*}

\begin{table*}
\caption{Individual Accuracy (\%) in terms of the three metrics for the 5-needle $M=1$ samples. We mark the best results with \textbf{bold face}. The index accuracy is not always $100$ \% because the model can fail to output the correct image index ``1''.}\label{tab:indiv_img1_needle5}
\vskip  0.3 cm
\centering
\resizebox{0.79\textwidth}{!}{%
\begin{tabular}{lcccccccccccc}
\toprule
Stitching & \multicolumn{2}{c}{$2\times2$} & \multicolumn{2}{c}{$4\times4$} & \multicolumn{2}{c}{$8\times8$}  \\
\cmidrule(r){1-1} \cmidrule(r){2-3} \cmidrule(r){4-5} \cmidrule(r){6-7} 
Metrics &   \emph{Index} & \emph{Exact}  & \emph{Index} & \emph{Exact}   & \emph{Index} & \emph{Exact}\\ 
\midrule
{API-based models}\\
\midrule
Claude 3 Opus &80.20&46.40&84.16&14.20&80.87&1.74 \\
Gemini Pro 1.0 &58.60&15.60&28.80&5.40&10.80&0.20\\
Gemini Pro 1.5& \bf{98.20}&\bf{76.55}&\bf{98.40}&27.80&\bf{95.40}&\bf{25.00}\\
GPT-4V &86.45&70.11&92.45&{45.10}&91.31&6.26 \\
GPT-4o& 88.34&71.72&94.83&\bf{50.43}&92.18&14.81\\
\midrule
Open-source models\\
\midrule
CogVLM-17B& 0.00 &0.00 & 0.00 &0.00& 2.55 &0.00\\
CogVLM2-LLaMA-3&  2.42&0.81 & 7.72 &0.00& 6.17 &0.00\\
Fuyu-8B & 80.00&0.00& 26.00&0.00& 10.00&0.00\\
mPLUG-Owl-v2 & 30.53&0.76&32.88&0.00&18.18&0.00\\
InstructBLIP-Vicuna-13B &  2.00 &0.00 & 5.56 &0.00& 5.77 &0.00\\
InstructBLIP-Flan-T5-XXL&  88.89 &0.00 &  60.48 &0.00& 62.50 &0.00\\
IDEFICS2-8B &  26.42 &4.88 &  51.85 &1.85& 82.00 &0.00\\
LLaVA-LLaMA-3  &  30.00 &8.00 &  30.00 &1.60& 54.60 &1.40\\

\bottomrule
\end{tabular}%
}
\vskip 0cm
\end{table*}

\begin{table*}[!t]
\caption{Individual Accuracy (\%) in terms of the three metrics for the 2-needle $M=10$ samples. We mark the best results with \textbf{bold face.}}\label{tab:indiv_img10_needle2}
\vskip  0.3 cm
\centering
\resizebox{0.79\textwidth}{!}{%
\begin{tabular}{lcccccccccccccc}
\toprule
Stitching & \multicolumn{2}{c}{$1\times1$} & \multicolumn{2}{c}{$2\times2$} & \multicolumn{2}{c}{$4\times4$} & \multicolumn{2}{c}{$8\times8$}  \\
\cmidrule(r){1-1} \cmidrule(r){2-3} \cmidrule(r){4-5} \cmidrule(r){6-7} \cmidrule(r){8-9} 
Metrics & \emph{Index} & \emph{Exact} & \emph{Index} & \emph{Exact}  & \emph{Index} & \emph{Exact} & \emph{Index} & \emph{Exact}\\ 
\midrule
{API-based models}\\
\midrule
Claude 3 Opus & 69.95&66.12&7.28&2.65&3.82&0.64&3.64&0.00\\
Gemini Pro 1.0 & 17.68&7.32&10.06&5.33&2.19&0.00&4.96&\bf{0.83}\\
Gemini Pro 1.5& {90.82}&{90.31}&{57.00}&{48.50}&18.32&{8.38}&2.53&0.00\\
GPT-4V &71.58&68.85&50.00&28.82&{22.42}&6.06&{11.18}&0.00\\
GPT-4o& \bf{94.82}&\bf{93.26}&\bf{88.89}&\bf{72.49}& \bf{40.59}&\bf{18.82}&\bf{19.21}&\bf{1.69}\\
\midrule
Open-source models\\
\midrule
Fuyu-8B & 4.00&0.00 &11.00&0.00&4.81&0.00&2.00&0.00\\
mPLUG-Owl-v2 & 1.68&0.84&1.64&0.00&7.69&0.00&4.55&0.00\\
IDEFICS2-8B & 3.00& 0.00& 4.95& 0.00& 0.00& 0.00 & 3.00& 0.00\\
LLaVA-LLaMA-3 & 0.00& 0.00& 1.00& 0.00& 0.00& 0.00 & 0.00& 0.00\\

\bottomrule
\end{tabular}%
}
\end{table*}

\begin{table*}[!t]
\caption{Individual Accuracy (\%) in terms of the three metrics for the 5-needle $M=10$ samples. We mark the best results with \textbf{bold face.}}\label{tab:indiv_img10_needle5}
\vskip  0.3 cm
\centering
\resizebox{0.79\textwidth}{!}{%
\begin{tabular}{lcccccccccccccc}
\toprule
Stitching & \multicolumn{2}{c}{$1\times1$} & \multicolumn{2}{c}{$2\times2$} & \multicolumn{2}{c}{$4\times4$} & \multicolumn{2}{c}{$8\times8$}  \\
\cmidrule(r){1-1} \cmidrule(r){2-3} \cmidrule(r){4-5} \cmidrule(r){6-7} \cmidrule(r){8-9} 
Metrics & \emph{Index} & \emph{Exact} & \emph{Index} & \emph{Exact}  & \emph{Index} & \emph{Exact} & \emph{Index} & \emph{Exact}\\ 
\midrule
{API-based models}\\
\midrule
Claude 3 Opus & 72.18&71.97&9.16&2.65&10.30&0.21&6.11&0.00
\\
Gemini Pro 1.0 & 21.20&12.00&12.40&3.00&11.16&0.44&7.44&0.00\\
Gemini Pro 1.5& {94.75}&{78.79}&{58.40}&{35.20}&{20.59}&{7.86}&{10.61}&0.41\\
GPT-4V &70.83&68.33&43.86&25.23&19.10&4.94&9.98&\bf{0.42}\\
GPT-4o& \bf{95.13}&\bf{91.81}&\bf{86.47}&\bf{56.14}&\bf{42.71}&\bf{19.89}&\bf{15.09}&0.26\\
\midrule
Open-source models\\
\midrule
Fuyu-8B & 14.00 &0.00 &9.00&0.00&13.00&0.00&8.00&0.00\\
mPLUG-Owl-v2 & 4.40&0.00&7.47&0.57&8.15&0.00&5.42&0.00\\
IDEFICS2-8B & 0.00& 0.00& 0.83& 0.00& 0.00& 0.00 & 0.00& 0.00\\
LLaVA-LLaMA-3 & 0.00& 0.00& 0.00& 0.00& 0.00& 0.00 & 0.00& 0.00\\

\bottomrule
\end{tabular}%
}
\vskip -0.5cm
\end{table*}

\end{document}